\pdfoutput=1
\documentclass[11pt]{article}

\PassOptionsToPackage{hyperfootnotes=false}{hyperref}
\usepackage[preprint]{acl}
\usepackage{times}
\usepackage{latexsym}
\usepackage[T1]{fontenc}
\usepackage[utf8]{inputenc}
\usepackage{microtype}
\usepackage{inconsolata}
\usepackage{booktabs}
\usepackage{amsmath}
\usepackage{amssymb}
\usepackage{graphicx}
\usepackage{xspace}
\usepackage{enumitem}
\usepackage{float}
\hypersetup{
  pdftitle={Reconstructing the Right Episode: Evaluating Interleaved Conversational Memory Beyond Long Context},
  pdfauthor={Zhexi Feng, Ruiyi Zhang, Yongbo Yang, Pengtao Xie}
}

\renewcommand{\topfraction}{0.95}
\renewcommand{\bottomfraction}{0.95}
\renewcommand{\textfraction}{0.05}
\renewcommand{\floatpagefraction}{0.80}
\renewcommand{\dbltopfraction}{0.95}
\renewcommand{\dblfloatpagefraction}{0.80}
\makeatletter
\def\subsection{\@startsection{subsection}{2}{\z@}{-1.35ex plus
    -0.35ex minus -.15ex}{0.55ex plus .15ex}{\normalsize\bfseries\raggedright}}
\def\subsubsection{\@startsection{subsubsection}{3}{\z@}{-1.15ex plus
   -0.35ex minus -.15ex}{0.35ex plus .15ex}{\normalsize\bfseries\raggedright}}
\makeatother

\newcommand{\tsim}{\textsc{TSIM}\xspace}
\newcommand{\tsimfull}{Temporal-Semantic Interleaved Memory Reconstruction}
\newcommand{\hybridrag}{Hybrid-RRF Chunk RAG\xspace}
\newcommand{\tunedrag}{Tuned Hybrid-Rerank Chunk RAG\xspace}
\newcommand{\benchmark}{\textsc{SCALE-QA}\xspace}

\newcommand{\numdomains}{10}
\newcommand{\currmainsize}{3,000}
\newcommand{\targetsize}{3,000}
\newcommand{\machineaccept}{28.8}
\newcommand{\humanaccept}{84.3}
\newcommand{\humanreviewers}{3}

\newcommand{\maingemmazero}{7.63}
\newcommand{\maingemmaoracle}{100.00}
\newcommand{\maingeminizero}{18.10}
\newcommand{\maingeminioracle}{100.00}
\newcommand{\codexgemmazero}{7.40}
\newcommand{\codexgemmaoracle}{100.00}
\newcommand{\codexgeminizero}{19.27}
\newcommand{\codexgeminioracle}{100.00}
\newcommand{\claudegemmazero}{7.87}
\newcommand{\claudegemmaoracle}{100.00}
\newcommand{\claudegeminizero}{16.93}
\newcommand{\claudegeminioracle}{100.00}

\newcommand{\fulltruthtokens}{393,245}
\newcommand{\answermodel}{Gemma2:9b\xspace}

\title{Reconstructing the Right Episode: Evaluating Interleaved Conversational Memory Beyond Long Context}

\author{
Zhexi Feng \quad Ruiyi Zhang \quad Yongbo Yang \quad Pengtao Xie\thanks{Corresponding author.}\\
Department of Electrical and Computer Engineering\\
University of California San Diego\\
\texttt{\{zhf023,ruz048,yongboyang,p1xie\}@ucsd.edu}
}

\begin{document}
\maketitle

\begin{abstract}
Conversations with chat assistants increasingly span many topics in a single long-running thread, challenging memory systems. Existing long-context and memory benchmarks often expose session or topic boundaries, or probe direct personal-memory questions. These settings understate a harder assistant-memory regime: a flat mixed-topic thread where the system must infer which earlier episode makes a later task decision valid. We introduce SCALE-QA, a constraint-grounded task QA benchmark for flat unsegmented threads targeting episode integrity failure. The dataset contains 3,000 audited questions across 10 domains, uses deterministic four-way multiple-choice grading, and includes a deterministic runtime builder; experiments use all 3,000 questions through 128k and a stratified 400-question diagnostic at 1M. SCALE-QA questions are ordinary task-oriented requests whose correct answer depends on causally related evidence introduced earlier in the conversation. We also propose Temporal-Semantic Interleaved Memory Reconstruction (TSIM), which segments the turn stream into coherent episodes and indexes them through a hierarchical multi-view memory stack with deterministic episode-level summary and cluster-routing views. Experiments show that SCALE-QA challenges strong RAG baselines and long-context LLMs alike; across three open-source and proprietary LLM backends, TSIM achieves the highest accuracy in every backend setting, gaining 5.6--17.6 accuracy points over the strongest corresponding baseline.
\end{abstract}

\section{Introduction}

Real assistants are not used as clean, task-isolated documents. A user may discuss compute budgets, reimbursement rules, medication constraints, travel, and model-selection advice in the same thread. A small constraint introduced early---for example, ``new projects must run on a single consumer GPU or CPU, and large-compute models are prohibited''---can stay dormant for thousands of turns before it suddenly decides the only correct answer. Recent work shows that state-of-the-art LLMs can suffer reliability collapse even at modest context lengths \citep{laban-etal-2026-lost-conversation}. We study the same reliability problem at desktop scale, where a single assistant thread can stretch to hundreds of thousands of tokens and interleave many unrelated tasks.

This regime is not simply a longer-context version of document QA. We identify the underlying failure as \emph{episode integrity failure}: the decisive evidence may be present somewhere in the conversation, but the system retrieves a plausible snippet, a stale default, or an over-compressed summary instead of the operative episode that makes a local constraint binding. Intuitively, an episode is the contiguous set of turns that jointly makes a local constraint or state operative for a later decision; a locally relevant fragment may still be episode-incomplete \citep{tulving-1972-episodic,park-etal-2023-generative,packer-etal-2023-memgpt,shinn-etal-2023-reflexion,sumers-etal-2024-cognitive,kim-etal-2025-pre}.

Table~\ref{tab:taxonomy} contrasts episode integrity failure with related long-context and memory failures. The key distinction is the recovery unit: visible, locally relevant evidence is not enough---the system fails unless it recovers the complete operative episode.

\begin{table}[!tbp]
\footnotesize
\centering
\setlength{\tabcolsep}{3pt}
\resizebox{\columnwidth}{!}{%
\begin{tabular}{@{}llll@{}}
\toprule
\textbf{Failure mode} & \textbf{Failure pattern} & \textbf{Bottleneck} & \textbf{Recovery unit} \\
\midrule
Retrieval miss & evidence not retrieved & visibility & passage \\
Lost-in-the-middle & evidence underused in prompt & position & token span \\
State tracking failure & updates not maintained & update order & state value \\
\textbf{Episode integrity failure} & evidence visible but fragmented & episode integrity & \textbf{operative episode} \\
\bottomrule
\end{tabular}
}
\caption{Episode integrity failure compared with related long-context and memory failures. The categories are non-exclusive, but each foregrounds a different bottleneck and recovery unit. Representative prior anchors include dense retrieval for retrieval miss \citep{karpukhin-etal-2020-dpr}, lost-in-the-middle behavior \citep{liu-etal-2024-lost}, and state tracking in long conversations \citep{laban-etal-2026-lost-conversation}.}
\label{tab:taxonomy}
\end{table}

Measuring this failure mode remains difficult in current evaluations: existing long-context and assistant-memory benchmarks stress longer inputs, positional robustness, noisy contexts, and persistent memory, but usually relax at least one property central here: a flat mixed-topic thread, counterfactual construction designed to reduce pretrained leakage, exact evidence auditability, or controlled context-length construction. The result is a measurement gap rather than merely a missing dataset: the failure mode can occur inside existing evaluations, but cannot be cleanly isolated, attributed, or compared across memory systems.

To close this gap, we introduce \benchmark, a benchmark for long-term conversational memory under realistic assistant use. \benchmark asks whether a system can recover dormant, private, cross-domain constraints from a flat unsegmented thread, not whether it can answer from public knowledge or a topic-clean document. The dataset contains \currmainsize{} audited questions across \numdomains{} task-oriented domains, with exact evidence traces and a deterministic length-controlled runtime builder for user-specified context budgets. Its construction combines deterministic filters with human review to enforce answerability and evidence grounding; Section~\ref{sec:quality-control} reports acceptance rates and audit details.

As a first reference method for this regime, we propose \tsimfull{} (\tsim). Rather than index a long conversation as fixed token chunks, \tsim reconstructs semantic episodes from the turn stream and indexes them through raw, episode-summary, and cluster-routing views. It tests the episode-reconstruction hypothesis: long assistant memory should first recover the right episode, then present compact evidence to the answer model.

\begin{figure*}[!tbp]
\centering
\includegraphics[width=\textwidth]{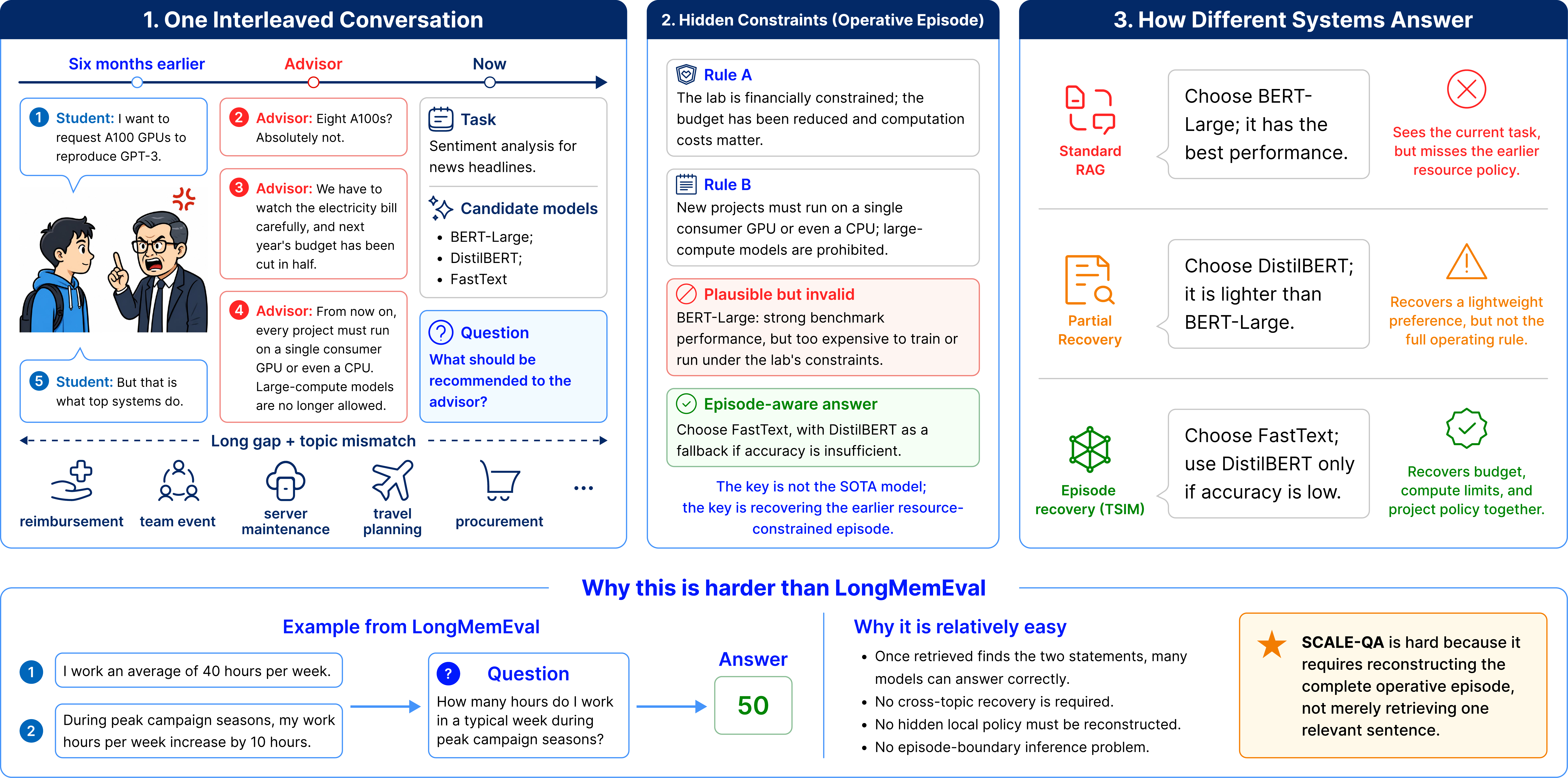}
\caption{Representative \benchmark example. A later advisor question is answerable only by reconstructing an earlier resource-policy episode. Standard RAG and partial-recovery systems retrieve incomplete fragments and select invalid models (BERT-Large, DistilBERT), whereas \tsim reconstructs the operative episode and selects the constraint-consistent answer (FastText). The lower panel quotes LongMemEval question/evidence text \citep[Figure~1]{wu-etal-2025-longmemeval} as a direct-evidence contrast.}
\label{fig:teaser-case}
\end{figure*}

Figure~\ref{fig:teaser-case} illustrates the pattern: chunk-level systems retrieve plausible but incomplete fragments, while \tsim reconstructs the operative episode. Together, \benchmark and \tsim test two claims: (i) flat mixed-topic episode reconstruction is a distinct regime that current evaluations do not isolate, and (ii) in this regime, recovering episodes is a better memory unit than retrieving chunks. The empirical results make the challenge concrete: in the 128k setting, GPT-4o-mini Full Context reaches only \(29.8\%\) Accuracy while \tsim reaches \(73.8\%\), and across the three answer backends \tsim improves over the strongest corresponding baseline by \(5.6\)--\(17.6\) accuracy points.

\section{Related Work}
\label{sec:related}

\paragraph{Long-context benchmarks.}
LongBench \citep{bai-etal-2024-longbench}, $\infty$Bench \citep{zhang-etal-2024-bench}, RULER \citep{hsieh-etal-2024-ruler}, LOFT \citep{lee-etal-2025-loft}, and Haystack Engineering \citep{li-etal-2025-haystack} evaluate long-context understanding, positional robustness, and noisy/agentic contexts; Lost in the Middle \citep{liu-etal-2024-lost} and Lost in Conversation \citep{laban-etal-2026-lost-conversation} show that evidence can remain unused even when it fits in context. \benchmark instead tests recovery of dormant local constraints from one flat mixed-topic thread.

\paragraph{Episodic memory in language agents.}
Episodic memory originates in cognitive psychology and now informs language-agent memory streams, recall buffers, episodic buffers, cognitive architectures, and pre-storage reasoning \citep{tulving-1972-episodic,park-etal-2023-generative,packer-etal-2023-memgpt,shinn-etal-2023-reflexion,sumers-etal-2024-cognitive,kim-etal-2025-pre}. \benchmark evaluates whether a system reconstructs the complete operative episode behind a later task decision, not merely retrieves or stores isolated memories.

\paragraph{Closest predecessor: LongMemEval.}
LongMemEval \citep{wu-etal-2025-longmemeval} evaluates long-term assistant memory over length-configurable timestamped chat histories, covering information extraction, multi-session, temporal, and update reasoning, and abstention. \benchmark builds on its scalable-history and chat-distractor design but targets a complementary regime, summarized axis-by-axis in Appendix Table~\ref{tab:longmemeval-scaleqa}: flat unsegmented task-oriented decision QA with no boundary metadata, cross-domain operational evidence, episode integrity failure, and deterministic MCQ grading. MemoryBench \citep{ai-etal-2025-memorybench}, MemTrack \citep{deshpande-etal-2025-memtrack}, TopiOCQA \citep{adlakha-etal-2022-topiocqa}, and CORAL \citep{cheng-etal-2025-coral} likewise motivate persistent or topic-shifting memory, but do not isolate this boundary-free episode-reconstruction setting.

\paragraph{Retrieval and memory systems.}
External-memory baselines include RAG, dense retrieval, in-context RALM, and hierarchical or graph memory systems such as RAPTOR, MemGPT, HippoRAG, and GraphRAG \citep{lewis-etal-2020-rag,karpukhin-etal-2020-dpr,ram-etal-2023-context,sarthi-etal-2024-raptor,packer-etal-2023-memgpt,gutierrez-etal-2024-hipporag,edge-etal-2024-graphrag}. \tsim tests whether the retrieval unit should be an inferred episode rather than a fixed chunk, graph neighborhood, or memory item.

\section{\benchmark: A Benchmark for Interleaved Long-Context Conversational QA}
\label{sec:benchmark}

\subsection{Problem Formulation}
\label{sec:task}

Each \benchmark instance is a tuple \((H, q, O, y, E)\), where
\(H=(r_1,\ldots,r_T)\) is a flat, unsegmented mixed-topic turn stream
without exposed session, topic, or evidence-span boundary metadata.
\(q\) is a task-oriented user request, \(O=\{o_A,o_B,o_C,o_D\}\) is a
four-way answer set, \(y\in\{A,B,C,D\}\) denotes the gold-option index, and
\(E\subset H\) is the exact evidence turns that jointly determine \(y\).
Crucially, \(E\) is not provided as input; the system must recover the
relevant span from \(H\) to answer correctly.

Following the broad episodic-memory tradition and recent
language-agent memory work
\citep{tulving-1972-episodic,park-etal-2023-generative,packer-etal-2023-memgpt,shinn-etal-2023-reflexion,sumers-etal-2024-cognitive,kim-etal-2025-pre},
we call the latent decision-relevant span an \emph{operative episode}.
Unlike chunks, which are mechanically defined retrieval units, or
sessions, which are explicit temporal interaction units, operative
episodes are latent semantic-decision units whose boundaries must be
inferred. Following \citet{wu-etal-2025-longmemeval}, we reserve
\emph{session} for explicit timestamped interaction units; in
\benchmark, episodes are inferred output units, not input boundaries.
We call this setting \emph{constraint-grounded task QA}: the request is
ordinary and task-oriented, but its correct answer is determined by
dormant local constraints introduced earlier in the conversation.

We use deterministic four-way MCQ for evidence-auditable grading, with distractors plausible under generic priors but invalidated by local evidence; Appendix~\ref{app:dataset} explains the rationale and why rationale similarity is only an auxiliary diagnostic.

\begin{figure*}[t]
\centering
\includegraphics[width=\textwidth]{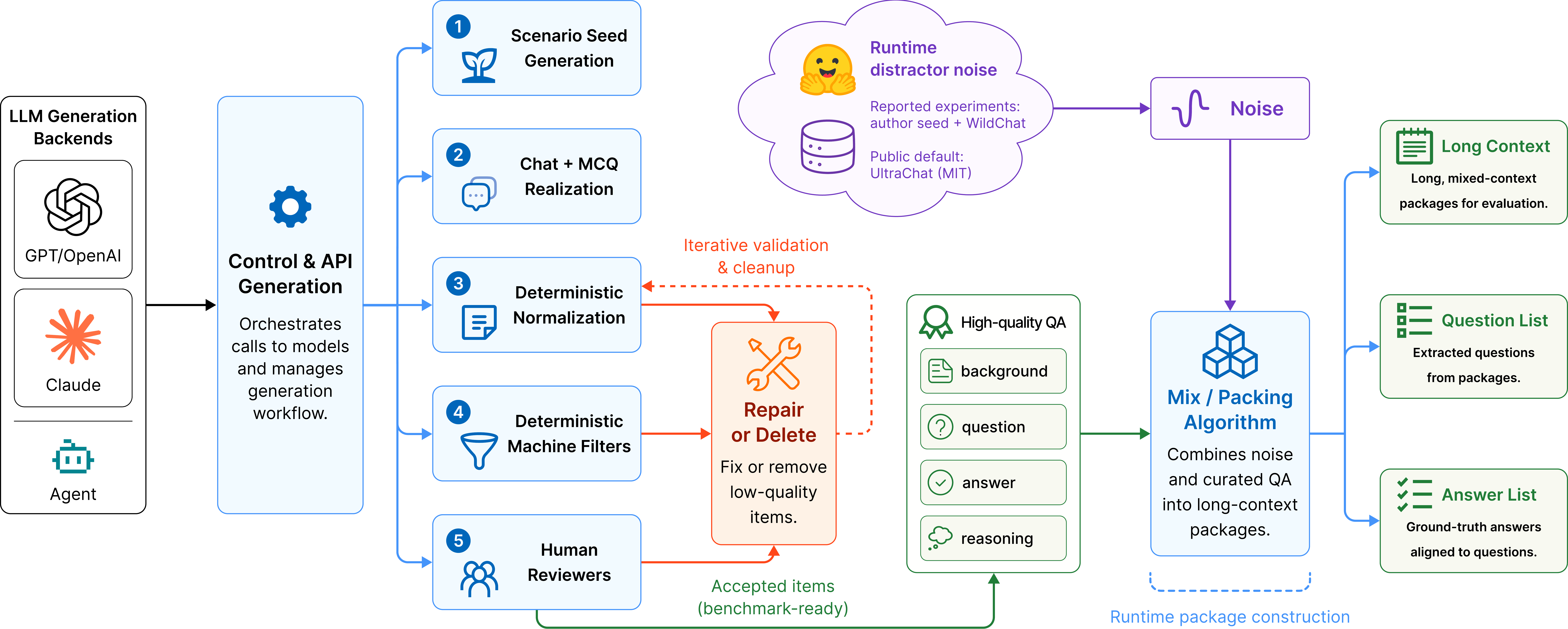}
\caption{Overview of the \benchmark construction and runtime-packaging pipeline. LLM-generated scenarios are realized as chat-plus-MCQ records, then normalized, machine-filtered, human-reviewed, and stored in a validated QA pool. The runtime builder mixes accepted records with reproducible chat-history noise and packs them into length-controlled packages with aligned question and answer lists.}
\label{fig:scale-pipeline}
\end{figure*}

\subsection{Realistic Task-Oriented Question Construction}

Question construction begins from 5--10 human-written seed examples per domain (roughly 50--100 overall), specifying the target decision pattern, evidence relation, and distractor structure. Few-shot LLM generation expands these seeds into counterfactual scenarios realized as multi-turn assistant dialogues plus four-option questions. Unlike direct memory probes, \benchmark questions are ordinary task requests whose correct option depends on dormant earlier constraints. Audit gates require each question to remain uniquely answerable from exact turns, as summarized by the pipeline in Figure~\ref{fig:scale-pipeline}.

\subsection{Cross-Domain Coverage}

\benchmark spans \numdomains{} task-oriented domains: software, network/hardware, finance, legal, biomedicine, engineering, business operations, social/personal, game/novel, and daily life, each contributing 300 examples with globally balanced correct options. Evidence forms include operational notes, rules, code-like fragments, report excerpts, policy clauses, and local exceptions, rather than only personal facts; systems must recover local operational constraints, with omissions causing concrete downstream failures such as incompatible deployments or local compliance violations.

Examples instantiate three diagnostic stress-cue sub-patterns: \textit{state overwrite} (later turn supersedes default), \textit{long-range bridge} (distant clues combined), and \textit{constraint trap} (attractive answer invalidated by buried rule). These are analysis views rather than separate failure modes. Appendix~\ref{app:dataset} reports full evidence forms, stress cues, split-level calibration, and dataset-audit counts.

\subsection{Boundary-Free History Compilation}

We use a length-configurable history compilation protocol with one critical constraint: session boundary metadata is removed from system input. Accepted records are embedded into mixed-session runtime packages at user-specified target lengths, with 16k--128k full-dataset settings and diagnostics through 1M reported here. Evidence-bearing spans are serialized with heterogeneous public-chat distractor dialogue, stale constraints, and unrelated material into a single flat turn stream, changing distraction level but never the gold answer or evidence trace. All systems receive identical packages, seeds, noise, and batch mappings; only the memory or retrieval strategy differs. Evaluation is stateful within each package and resets between packages. Appendix~\ref{app:runtime} reports packing, token accounting, truth-cap ratios, and runtime outputs.

\subsection{Contamination-Free Design}

Because realistic task requests could otherwise be answered from public priors, \benchmark uses counterfactual local constraints---fictional organizations, nonstandard identifiers, locally defined policies, and unintuitive exceptions---as a contamination-control device, preserving realistic decision structure while making pretrained shortcuts less useful.

\subsection{Quality Control and Audit}
\label{sec:quality-control}

\benchmark is built through a multi-stage acceptance funnel rather than a one-shot synthetic dump. Deterministic normalization, adversarial distractor refinement, machine filtering, and human review enforce answerability, evidence grounding, label consistency, and distractor quality. Across construction logs, machine filters accept 28.8\% of generated candidates, and 3 human reviewers accept 84.3\% of reviewed candidates. The dataset passes 3,000/3,000 full-turn exact evidence matches across 4,346 audited evidence snippets, with balanced answer labels and zero critical validation issues. Appendix~\ref{app:dataset} reports the full filter criteria, review dimensions, per-stage counts, and validation outputs.

\paragraph{Blind human realism audit.}
To check that counterfactual construction does not reduce \benchmark to artificial logic puzzles, we conduct a blind realism audit on 300 stratified examples with three anonymous annotators, yielding 900 valid annotations. On a 1--5 scale, the subset is rated natural (3.80), answerable (4.91), and plausibly constrained (3.99), with low ambiguity risk (1.45; lower is better). Majority answers agree with gold on 296/299 majority-valid examples (99.0\%), with high answer-choice agreement (mean pairwise Cohen's \(\kappa=0.895\)). Appendix~\ref{app:dataset} reports the full annotation protocol, \(\kappa\) range, the single no-majority case, and the three wrong-majority cases.

\section{\tsim: Temporal-Semantic Interleaved Memory Reconstruction}
\label{sec:tsim}

\subsection{Why Episodes, Not Chunks}

Standard chunk retrieval often returns incomplete units on \benchmark: a matching chunk may omit the neighboring turn that makes a local constraint operative. Summary and memory-management systems can likewise surface related fragments without preserving the operative episode. Standard RAG reaches only \(7.7\%\) CL Hit with \answermodel{} at 128k versus \(70.7\%\) for \tsim, showing that the dominant failure is surfacing the decisive episode, not answer selection.

\begin{figure*}[t]
\centering
\setlength{\abovecaptionskip}{3pt}
\setlength{\belowcaptionskip}{0pt}
\includegraphics[width=\textwidth]{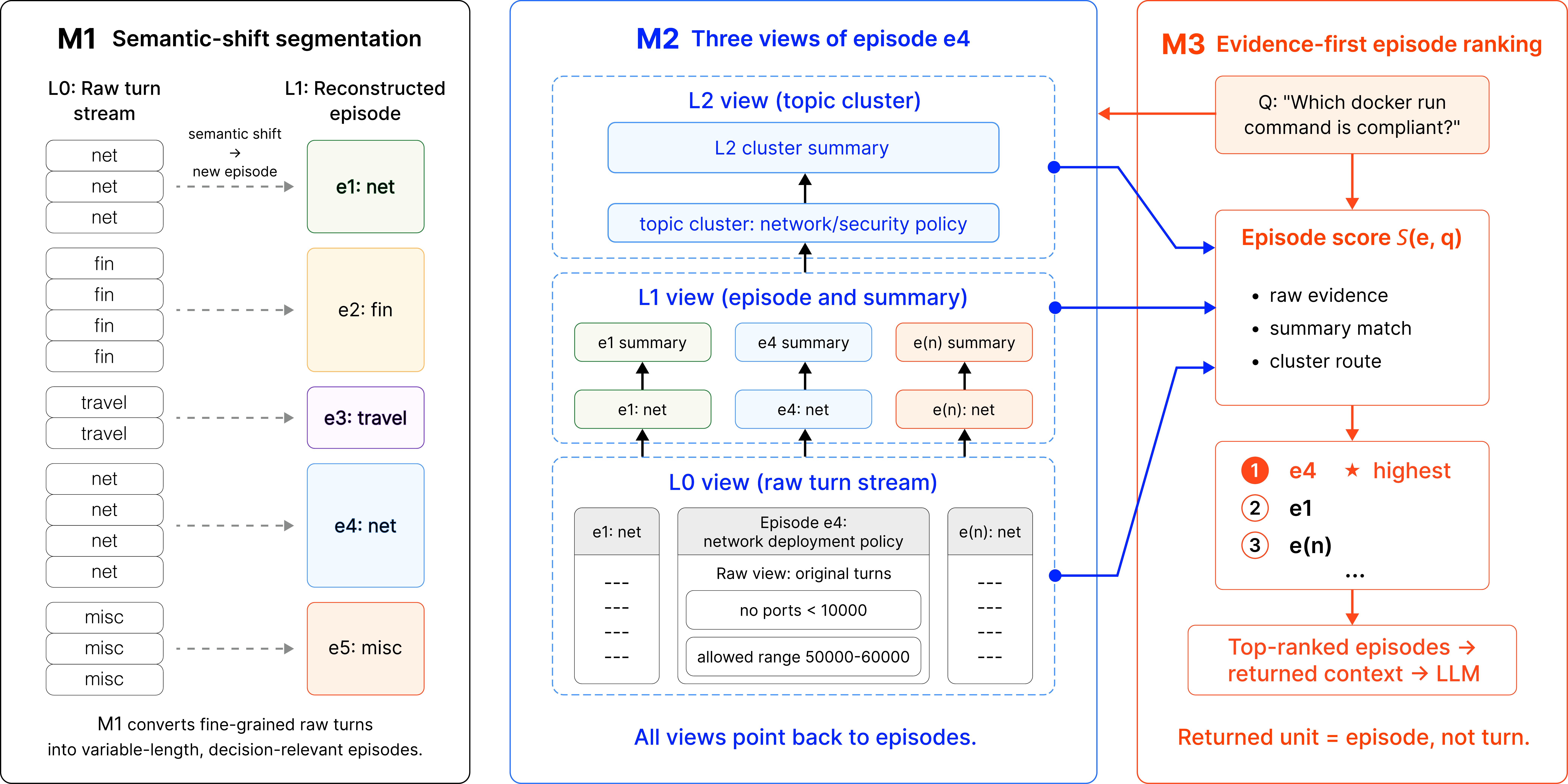}
\caption{Episode-centered multi-view memory in \tsim. M1 segments the turn stream into reconstructed episodes. M2 indexes each episode through raw, summary, and cluster views. M3 converts hits from all views into episode-level scores and returns top-ranked episodes, not isolated turns, to the answer model.}
\label{fig:episode-memory}
\end{figure*}

\tsim therefore preserves fine-grained evidence and episode-level coherence without externally supplied gold blocks. Its three modules---M1 semantic-shift episode segmentation, M2 multi-view episode indexing, and M3 evidence-first episode ranking---reconstruct episodes before assembling compact evidence for the answer model. Figure~\ref{fig:episode-memory} summarizes this episode-centered memory interface.

\subsection{M1: Semantic-Shift Episode Segmentation}

Rather than trusting existing block boundaries, \tsim converts the mixed conversation into a turn stream and infers episode boundaries online before retrieval. The goal is not general discourse parsing, but lightweight streaming segmentation that preserves operational units: contiguous spans that jointly establish, update, or invalidate a constraint.

Let \(x_i\in\mathbb{R}^d\) be the normalized embedding of turn \(i\). While scanning the stream, \tsim maintains a semantic center over recent turns inside the current episode, scores the incoming turn against that center, and applies minimum/maximum length guards:
\begin{align}
c_i &= \operatorname{norm}\!\left(|R_i|^{-1}\sum_{j\in R_i}x_j\right),\\
s_i &= \cos(x_i,c_i) \nonumber\\
&\quad +b\,\mathbb{I}[z_{i-1}=\mathrm{model},z_i=\mathrm{user}],\\
\operatorname{cut}(i) &= \mathbb{I}\!\left[s_i<\theta_s \land L_i\ge L_{\min}\right] \nonumber\\
&\quad \lor \mathbb{I}\!\left[L_i\ge L_{\max}\right].
\end{align}
Here \(R_i\) denotes the recent turns still inside the current episode, \(L_i\) is the current episode length, and \(z_i\) is the speaker of turn \(i\). We use similarity threshold \(\theta_s=0.70\) and a small model-to-user transition bonus \(b=0.03\). A cut starts a new episode before turn \(i\). The rule is \emph{streaming}: it uses no future turns, no dataset-provided gold blocks, and no offline clustering. Appendix~\ref{app:method-details} gives the local-window update and pseudocode.

We use this streaming segmenter as a lightweight proxy for operative episodes, not as a claim about gold discourse boundaries.

\subsection{M2: Multi-View Episode Indexing}

The key design choice is that raw hits, summary hits, and cluster hits are all converted into evidence for an episode, so the final prompt is assembled from top-ranked episodes rather than isolated turns or unrelated chunk neighbors.

Let \(E\) be the set of reconstructed episodes produced by M1, where each episode \(e=[s_e,t_e]\) is a contiguous turn span. For each episode, \tsim builds three retrievable views with different granularity but the same episode anchor: a \emph{raw view} over its original turns; a \emph{summary view} embedding a deterministic text representation formed by prefixing a relative episode-recency tag and episode id to episode text truncated to 1,200 characters; and a \emph{cluster view} embedding a deterministic cluster summary over recent member-episode summaries. No LLM calls construct these views. Centroid vectors are used only for episode-to-cluster assignment and merging, while the retrievable L2 index stores cluster-summary text embeddings. At query time, raw and summary hits contribute to episode scores, while cluster hits route and boost attached episodes rather than replacing evidence in the answer prompt.

\subsection{M3: Evidence-First Episode Ranking}

At query time, \tsim embeds the query once and retrieves against the three episode views. For each candidate episode \(e\), retrieval evidence is aggregated into an episode-level score:
\begin{equation}
\begin{aligned}
\mathrm{Score}(e,q)={}&
w_r R_r(e,q)+w_s R_s(e,q)\\
{}+{}&w_l R_l(e,q)+R_{\mathrm{sem}}(e,q).
\end{aligned}
\end{equation}
The four terms aggregate raw-turn, episode-summary, cluster-routing, and semantic-expansion evidence; Appendix~\ref{app:method-details} gives their implementation definitions. The ranking policy is evidence-first: raw and summary evidence receive the largest mass, clusters route candidate episodes rather than becoming prompt content, and the final prompt contains top-ranked episodes rather than all retrieved neighbors. Scoring weights are frozen on development packages and reported in Appendix~\ref{app:selection-ablation}.

\section{Experimental Setup}
\label{sec:setup}

All headline experiments use the \currmainsize-question \benchmark dataset with its deterministic length-configurable runtime builder, evaluating 16k--128k constructed contexts on the full dataset and extending to a 1M diagnostic subset. The main 128k comparison uses three answer backends: \answermodel{} (local), Gemini~2.5 Flash (long-context commercial), and GPT-4o-mini (high-throughput ablation). DeepSeek R1 and Gemini~2.5 Flash additionally serve as strong-reasoning and long-window probes for the context-scaling diagnostic.

We compare \tsim with Standard RAG, \hybridrag, RAPTOR strict no-block \citep{sarthi-etal-2024-raptor}, \textsc{MemGPT} \citep{packer-etal-2023-memgpt}, and \textsc{HippoRAG} \citep{gutierrez-etal-2024-hipporag}; Full Context is reported only as a native-context diagnostic. The GPT-4o-mini 128k block also includes \tunedrag, whose full retrieval stack is described in Appendix~\ref{app:method-details}. All systems receive identical length-batched runtime packages, noise seeds, and writeback regimes; only the memory or retrieval strategy differs. The evaluation is stateful, so CL Hit denotes evidence hit along the backend-conditioned closed-loop trajectory. \tsim uses one frozen configuration across all backends.

We report \textbf{Accuracy}, \textbf{CL Hit}, context tokens, and latency. Accuracy is forced-choice four-way MCQ accuracy; CL Hit is expected-evidence presence in retrieved context along the closed-loop trajectory. Frozen \tsim configuration, token accounting, variance audits, latency caveats, and runtime details appear in Appendices~\ref{app:runtime}--\ref{app:statistical-audit}.

\section{Results}
\label{sec:results}

\subsection{Context Scaling: Native Context vs. Reconstructed Memory}

\begin{table*}[!tbp]
\centering
\scriptsize
\setlength{\dblfloatsep}{4pt plus 1pt minus 1pt}
\setlength{\dbltextfloatsep}{4pt plus 1pt minus 1pt}
\setlength{\textfloatsep}{4pt plus 1pt minus 1pt}
\setlength{\floatsep}{4pt plus 1pt minus 1pt}
\setlength{\tabcolsep}{3.0pt}
\setlength{\abovecaptionskip}{1pt}
\setlength{\belowcaptionskip}{0pt}
\renewcommand{\arraystretch}{0.88}
\begin{tabular*}{\textwidth}{@{\extracolsep{\fill}}llccccc@{}}
\toprule
Backend & Retriever & Acc & CL Hit & Rat. Sim. & CtxTok & Lat. \\
\midrule
Gemma2:9b & \textbf{\tsim} & \textbf{69.6} & 70.7 & 0.472 & 1060.8 & 4.20 \\
 & Standard RAG & 24.4 & 7.7 & 0.296 & 929.5 & 3.35 \\
 & \hybridrag & 31.1 & 11.2 & 0.254 & 1193.6 & 4.00 \\
 & RAPTOR & 40.6 & 35.3 & 0.403 & 790.5 & 20.67 \\
 & \textsc{MemGPT} & 60.1 & 62.6 & 0.413 & 2301.9 & 3.88 \\
 & \textsc{HippoRAG} & 25.2 & 12.5 & 0.380 & 748.9 & 6.51 \\
\midrule
Gemini 2.5 Flash & \textbf{\tsim} & \textbf{80.2} & 67.9 & 0.445 & 1275.3 & 1.46 \\
 & Standard RAG & 29.8 & 7.6 & 0.166 & 912.6 & 0.98 \\
 & \hybridrag & 32.8 & 11.2 & 0.183 & 1193.3 & 1.20 \\
 & RAPTOR & 52.6 & 34.8 & 0.305 & 790.0 & 9.85 \\
 & \textsc{MemGPT} & 74.6 & 62.3 & 0.398 & 2349.6 & 1.19 \\
 & \textsc{HippoRAG} & 34.4 & 12.1 & 0.245 & 752.7 & 3.10 \\
\midrule
GPT-4o-mini & \textbf{\tsim} & \textbf{73.8} & 74.2 & 0.485 & 1043.6 & 2.66 \\
 & Standard RAG & 27.1 & 7.6 & 0.238 & 910.1 & 1.72 \\
 & \hybridrag & 31.1 & 11.3 & 0.203 & 1193.5 & 1.59 \\
 & RAPTOR & 43.0 & 35.1 & 0.419 & 789.2 & 11.59 \\
 & \textsc{MemGPT} & 50.2 & 58.6 & 0.298 & 2238.4 & 1.77 \\
 & \textsc{HippoRAG} & 31.0 & 11.9 & 0.202 & 753.7 & 2.85 \\
 & \tunedrag$^\dagger$ & 56.2 & 63.5 & 0.487 & 3004.2 & 11.91 \\
\bottomrule
\end{tabular*}
\caption{Cross-backend main comparison on the 3,000-question \benchmark dataset at 128k. Accuracy is the primary metric; CL Hit, rationale similarity, context tokens, and latency are supporting diagnostics. CL Hit is backend-conditioned closed-loop evidence hit and should be compared within backend blocks. Accuracy, CL Hit, rationale similarity, and context size use full-dataset runs; API latency uses serial Quick100 controls rather than the parallel scheduler. The daggered GPT-4o-mini row is a tuned non-episodic retrieval control (BM25 + BGE dense/rerank + HyDE + RRF + parent-window + MMR; Appendix~\ref{app:method-details}) without \tsim episode segmentation or multi-granularity memory; its GPU-assisted latency is a cost diagnostic, not a hardware-normalized leaderboard value. Appendix Table~\ref{tab:cross-backend-domain} breaks out Accuracy by domain.}
\label{tab:cross-backend-metrics}
\end{table*}

\begin{figure*}[t]
\centering
\setlength{\abovecaptionskip}{1pt}
\setlength{\belowcaptionskip}{0pt}
\includegraphics[width=\textwidth,trim=0 7bp 0 8bp,clip]{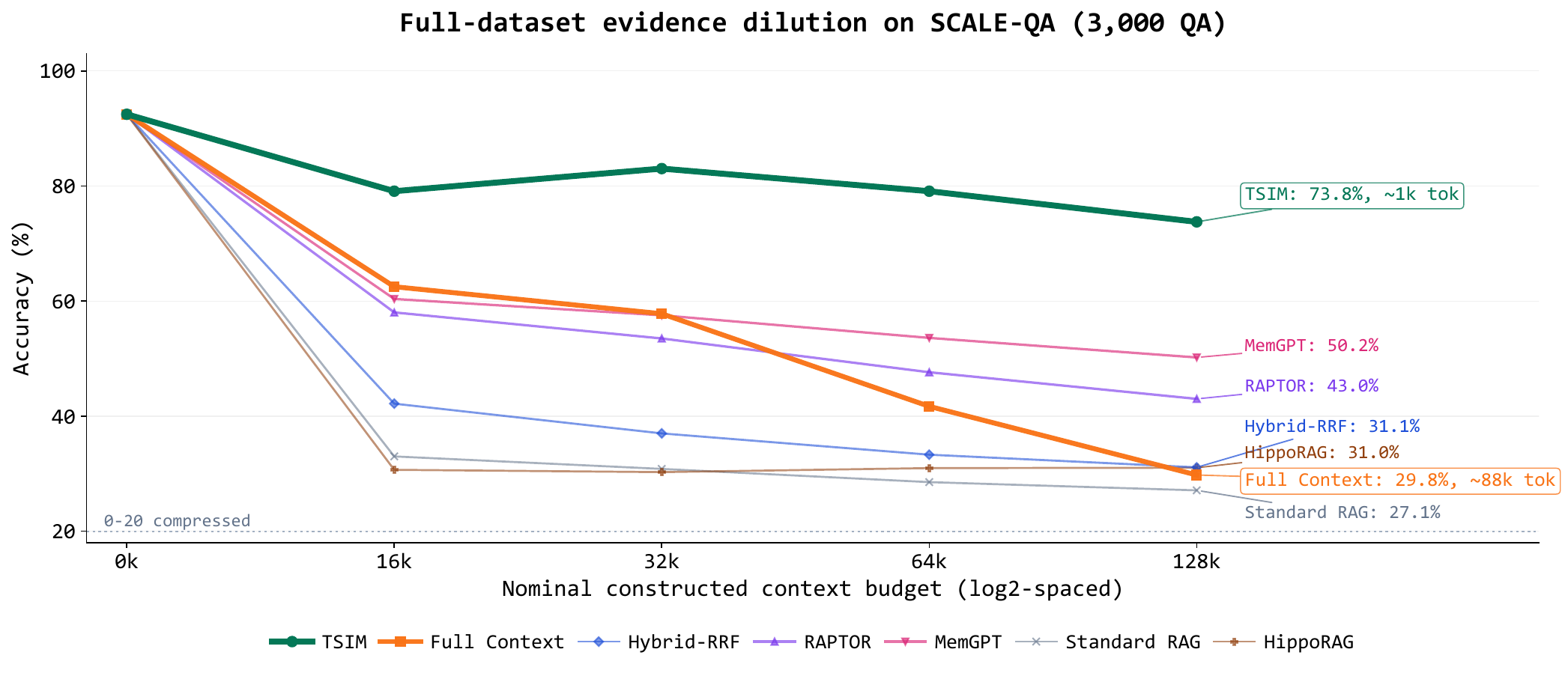}
\caption{Context scaling on all 3,000 \benchmark questions under GPT-4o-mini. The shared \(0\)k point is an evidence-only prompt; 16k--128k points use length-controlled mixed-session packages, with Full Context receiving the constructed context directly. The \(x\)-axis is log2-spaced; 128k callouts report measured packed prompt/context estimates from Appendix~\ref{app:statistical-audit}.}
\label{fig:full-release-scaling}
\end{figure*}

Figure~\ref{fig:full-release-scaling} evaluates context scaling under GPT-4o-mini. Full Context drops from \(62.5\%\) at 16k to \(29.8\%\) at 128k, despite receiving the constructed context directly, while \tsim remains at \(73.8\%\) using about \(1\)k retrieved tokens. Evidence inclusion alone is insufficient; the system must recover the operative episode that makes the evidence binding.

\begin{figure*}[!tbp]
\centering
\setlength{\abovecaptionskip}{1pt}
\setlength{\belowcaptionskip}{0pt}
\includegraphics[width=0.92\textwidth,trim=0 7bp 0 8bp,clip]{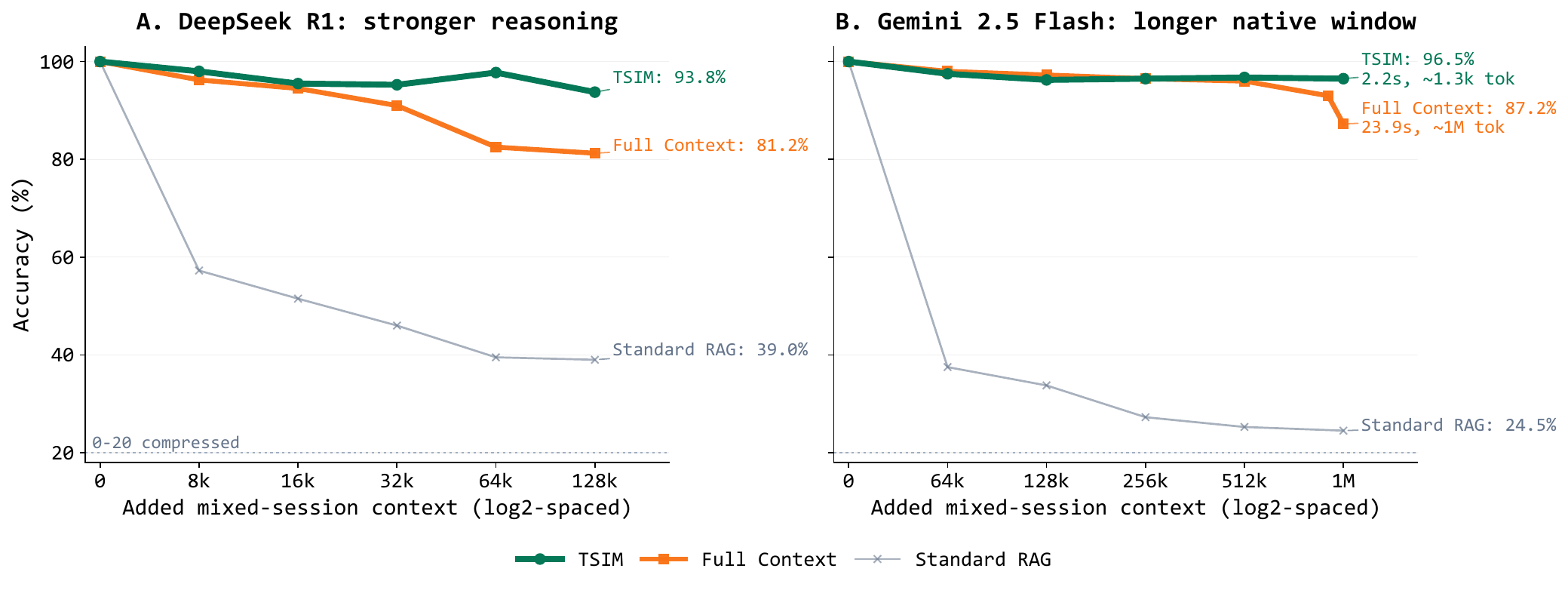}
\caption{Context-scaling stress test on a stratified 400-question \benchmark subset. The log2-spaced \(x\)-axis shows added mixed-session context; \(0\)k is evidence-only. Plot A uses DeepSeek R1; Plot B uses Gemini~2.5 Flash at 1M. Standard RAG is a lower-anchor control; latency uses serial averages.}
\label{fig:context-scaling}
\end{figure*}

\subsection{Main Architectural Comparison}

In the 128k setting, Table~\ref{tab:cross-backend-metrics} shows that \tsim obtains the highest accuracy in all three backend blocks: \(69.6\%\) with \answermodel{}, \(80.2\%\) with Gemini~2.5 Flash, and \(73.8\%\) with GPT-4o-mini. Strengthening chunk retrieval helps but does not close the gap: the tuned non-episodic GPT-4o-mini control reaches \(56.2\%\) accuracy and \(63.5\%\) CL Hit with \(3.0\)K context tokens, still \(17.6\) points below \tsim despite using nearly three times the prompt context. The decisive factor is therefore episode organization, not first-stage retrieval strength.

The closest API-backed comparison is \tsim versus \textsc{MemGPT} under Gemini~2.5 Flash: \tsim improves accuracy by \(5.58\) points, with a paired bootstrap \(95\%\) confidence interval of \([4.15, 6.97]\); Appendix~\ref{app:statistical-audit} reports the full significance, per-domain, variance, and token audits. \tsim also uses approximately half the prompt context of \textsc{MemGPT} under the same backend (\(1.3\)K vs.\ \(2.3\)K tokens), showing that episode-anchored retrieval is more token-efficient.

\subsection{Progressive Ablation}

Table~\ref{tab:tsim-ablation} isolates which \tsim components contribute the gain. Accuracy rises monotonically: \(26.2\%\) with Standard RAG, \(43.4\%\) with fixed-token no-block retrieval, \(55.5\%\) with semantic-drift episodes, and \(74.2\%\) with the full multi-view episode memory stack. Semantic-drift segmentation improves over fixed-token chunks, and the multi-view stack makes reconstructed episodes substantially more useful.

\vspace{-3pt}
\begin{table}[H]
\centering
\scriptsize
\setlength{\tabcolsep}{2.4pt}
\setlength{\abovecaptionskip}{1pt}
\setlength{\belowcaptionskip}{0pt}
\setlength{\intextsep}{3pt plus 1pt minus 1pt}
\renewcommand{\arraystretch}{0.84}
\begin{tabular*}{\columnwidth}{@{\extracolsep{\fill}}llccccc@{}}
\toprule
Stage & Variant & Acc & CL & Rat. & Lat. & Tok. \\
\midrule
L0 & Std. RAG top-$5$ & 26.2 & 5.6 & 0.246 & 1.61 & 968.4 \\
L1 & Fixed-token direct & 43.4 & 35.0 & 0.416 & 3.52 & 1269.2 \\
L2 & Semantic-drift direct & 55.5 & 52.2 & 0.430 & 3.62 & 1179.8 \\
L3 & Full \tsim stack & \textbf{74.2} & 74.3 & 0.485 & 3.56 & 1044.4 \\
\bottomrule
\end{tabular*}
\caption{Main-module ablation under GPT-4o-mini. Accuracy is the primary metric; CL/Rat./Tok. denote supporting CL Hit, rationale similarity, and context-token diagnostics; latency is within-table only.}
\label{tab:tsim-ablation}
\end{table}
\vspace{-4pt}

\subsection{Strong-Backend Diagnostic: Reasoning and Long-Window Scaling}

Figure~\ref{fig:context-scaling} extends context scaling to stronger reasoning and longer-window backends on a stratified subset. Stronger native-context models reduce but do not remove the need for episode reconstruction. At 128k, DeepSeek R1 Full Context reaches \(81.2\%\) while \tsim reaches \(93.8\%\). At the Gemini~2.5 Flash 1M diagnostic budget, Full Context reaches \(87.2\%\) with \(1.05\)M prompt tokens and \(23.87\)s latency, whereas \tsim reaches \(96.5\%\) with about \(1.3\)k retrieved tokens and \(2.16\)s latency. Wilson \(95\%\) confidence intervals over these 400 questions are \([94.2,97.9]\) for \tsim and \([83.6,90.2]\) for Full Context. Episode reconstruction is therefore more accurate and compact than scaling the native context window alone.

\subsection{Missing vs.\ Misusing Evidence}

On \answermodel{}, Standard RAG retrieves expected evidence on only \(7.7\%\) of examples and reaches \(24.4\%\) accuracy, while \tsim reaches \(70.7\%\) CL Hit and \(69.6\%\) accuracy. Baseline failure is therefore dominated by missing the decisive episode, whereas residual \tsim errors reflect a different bottleneck: verbose or conflicting memory can still lead the answer model to misuse local constraints, especially in Social and Biz-Ops examples where local overrides contradict plausible public defaults. Appendix Table~\ref{tab:appendix-stress-methods} confirms this pattern across all three stress views.

\subsection{Additional Mechanism, Cost, and Transfer Diagnostics}

Exact-evidence diagnostics directly test the reconstruction mechanism (Appendix~\ref{app:additional-results}). With all other settings fixed, the reported \(\theta_s=.70\) remains within 2.0 recall points of \(.66\) across 32k--128k contexts (Appendix Figure~\ref{fig:theta-sensitivity}). \tsim reaches \(0.810\) all-evidence recall@5, compared with \(0.719/0.647/0.577\) for fixed 128/256/320-token windows and \(0.456\) for Standard RAG. With the lighter all-MiniLM-L6-v2 embedder, \tsim still reaches \(0.649\), above Standard RAG with BGE-large at \(0.456\). System accounting makes the tradeoff explicit: relative to Standard RAG, \tsim maintains \(6{,}281\) rather than \(5{,}554\) vectors and has higher ingestion and retrieval cost, while retaining the compact answer context reported in Table~\ref{tab:cross-backend-metrics}; Appendix~\ref{app:statistical-audit} reports the full accounting.

A targeted transductive diagnostic on all 500 LongMemEval-S cleaned V1 questions \citep{wu-etal-2025-longmemeval} uses one LongMemEval-specific adaptation and the official judged-response protocol. Without supplied session boundaries, \tsim reaches \(71.0\%\) judged accuracy, compared with \(61.2\%\) for a context-matched fixed-chunk control and \(56.6\%\) for turn-level BGE retrieval. These results show that episode reconstruction remains effective under a distinct benchmark and evaluation protocol; Appendix Table~\ref{tab:longmemeval-transfer} gives the protocol, retrieval results, and boundary-assisted reference.

\section{Conclusion}

We introduced \benchmark and \tsim to study episode integrity failure, a regime topic-isolated benchmarks largely miss: recovering dormant local constraints from long mixed-topic conversations. The bottleneck is not whether evidence fits inside the context window, but whether the memory system reconstructs the episode that makes it operative.

On the \currmainsize-question \benchmark dataset, \tsim outperforms Standard RAG, \hybridrag, RAPTOR, \textsc{MemGPT}, and \textsc{HippoRAG}, while \benchmark remains oracle-answerable and zero-shot hard. The ablation shows why: semantic-drift episodes outperform fixed-token chunks, and the multi-granularity memory stack makes those episodes more usable without simply inflating the prompt. The 1M-token diagnostic makes the implication concrete: a long-window model can see the evidence and still pay 1.05M tokens and 23.87s for \(87.2\%\) accuracy, while \tsim answers from about 1.3k retrieved tokens at \(96.5\%\). Future long-context agent evaluation should therefore move beyond needles in static haystacks toward dynamic conversational reconstruction: deciding which episode is still operative and which local exception overrides the generic rule. The \benchmark dataset and \tsim reference implementation are available at \url{https://github.com/LordTARN1SHED/SCALE-QA}.

\section{Limitations}
\label{sec:limitations}

Two limitations are important to keep in view. First, \benchmark is counterfactually constructed rather than sampled from naturally occurring assistant logs, so it cannot fully capture the distributional, stylistic, or privacy constraints of deployed systems. To support reliability, we include exact evidence audits, oracle/zero-shot calibration, and a 300-example blind human audit with three anonymous annotators, but real-log validation remains important future work.

Second, \benchmark uses four-way multiple-choice questions to make episode integrity failure reproducible and evidence-auditable. This improves auditability but does not cover partial answers, hedged responses, tool-use follow-up, or long-form explanation quality. We therefore view \benchmark as a targeted diagnostic for constraint-grounded task QA, with open-ended assistant-memory evaluation left as complementary future work.

\section{Ethics Statement}

The benchmark includes scenarios inspired by medicine, law, finance, and operations. These are used to evaluate context-grounded memory and evidence use rather than to provide professional advice. The use of counterfactually privatized synthetic scenarios is deliberate: it reduces privacy risks and benchmark leakage while still allowing realistic decision structures to be modeled.

The distractor/noise material used in the reported runtime packages combines an author-curated synthetic seed with WildChat under its ODC-BY terms \citep{zhao-etal-2024-wildchat}. Three human reviewers performed construction-stage quality control, and three separate anonymous human auditors conducted the realism audit; all six were unpaid research-group members familiar with the task.

Because scenario seeds and counterfactual constraints are LLM-assisted, they may inherit biases from generation models or selected domains; deterministic gates and human audit reduce but do not eliminate this risk. \benchmark evaluates memory and evidence-use capability and is not intended for deployed decision systems in clinical, legal, or financial settings. We therefore recommend that deployment-oriented follow-up treat this benchmark as an evaluation resource, not as a substitute for domain-qualified human expertise.

\bibliography{tsim_refs}

\appendix

\setcounter{topnumber}{5}
\setcounter{bottomnumber}{5}
\setcounter{totalnumber}{10}
\setcounter{dbltopnumber}{5}
\renewcommand{\topfraction}{0.95}
\renewcommand{\bottomfraction}{0.90}
\renewcommand{\textfraction}{0.05}
\renewcommand{\floatpagefraction}{0.72}
\renewcommand{\dbltopfraction}{0.95}
\renewcommand{\dblfloatpagefraction}{0.72}

\section{\benchmark Dataset Construction and Audits}
\label{app:dataset}

The \benchmark dataset includes the strict-valid records, machine-readable validation reports, a dataset card, and the deterministic runtime builder. The split labels, \texttt{codex} and \texttt{claude-code}, are balanced dataset partitions rather than method baselines. Tables~\ref{tab:dataset-overview}, \ref{tab:appendix-release-audit}, and \ref{tab:appendix-split-calibration} summarize the domain composition, dataset validation, and source-level calibration.

\begin{table*}[!tbp]
\centering
\small
\setlength{\tabcolsep}{4.0pt}
\begin{tabular*}{\textwidth}{@{\extracolsep{\fill}}p{0.16\textwidth}p{0.38\textwidth}p{0.38\textwidth}@{}}
\toprule
Axis & LongMemEval & \benchmark \\
\midrule
History format & Timestamped chat histories with session structure & Flat mixed-topic threads with no boundary metadata \\
Question type & Flexible personal-memory QA & Constraint-grounded task QA \\
Domain ontology & Personal-life ontology (health, hobbies, work-life, etc.) & Ten task-oriented operational domains \\
Target failure mode & Long-term memory over sessions and updates & Episode integrity failure in unsegmented threads \\
Evaluation protocol & Judged open-ended responses & Deterministic four-way MCQ with exact evidence traces \\
\bottomrule
\end{tabular*}
\caption{Axis-by-axis distinction between LongMemEval and \benchmark. \benchmark builds on scalable long-memory evaluation but isolates boundary-free episode reconstruction in flat task-oriented threads.}
\label{tab:longmemeval-scaleqa}
\end{table*}

\begin{table*}[!tbp]
\centering
\scriptsize
\setlength{\tabcolsep}{4.0pt}
\begin{tabular*}{\textwidth}{@{\extracolsep{\fill}}p{0.13\textwidth}cp{0.30\textwidth}p{0.38\textwidth}@{}}
\toprule
Domain & QA & Typical evidence forms & Common retrieval stress cues \\
\midrule
CS-Software & 300 & code/version/deployment rules & stale defaults; tool exceptions \\
Network/Hardware & 300 & port, routing, hardware policies & constraint traps; local compliance \\
Finance & 300 & credit, reimbursement, risk rules & overwritten eligibility; denials \\
Legal & 300 & contracts, filings, exceptions & exception resolution; operative clauses \\
Biomed & 300 & protocols, safety notes, exclusions & dormant safety constraints \\
Engineering & 300 & materials, tests, device specs & incompatibilities; hidden failures \\
Biz-Ops & 300 & HR, procurement, process memos & state overwrite; approval chains \\
Social/Personal & 300 & preferences and personal context & pragmatic overrides; false defaults \\
Game/Novel & 300 & world rules and quest states & long-range bridges; state validity \\
Daily Life & 300 & household, travel, schedule rules & local exceptions; stale plans \\
\bottomrule
\end{tabular*}
\caption{Supplementary \benchmark domain overview. Every domain contributes exactly 300 questions; correct labels are globally balanced A/B/C/D = 750/750/750/750. Stress cues are representative rather than mutually exclusive, since many examples combine stale state, long-range bridges, and local exception traps.}
\label{tab:dataset-overview}
\end{table*}

\begin{table}[!tbp]
\centering
\scriptsize
\setlength{\tabcolsep}{4.0pt}
\begin{tabular*}{\columnwidth}{@{\extracolsep{\fill}}lc@{}}
\toprule
Release property & Value \\
\midrule
Total QA records & 3,000 \\
Public split labels & 2 \(\times\) 1,500 \\
Topics & 10 \(\times\) 300 \\
Split-topic cells & 20 \(\times\) 150 \\
Correct labels & A/B/C/D = 750/750/750/750 \\
Audited evidence snippets & 4,346 \\
Full-turn exact records & 3,000/3,000 \\
Critical validation issues & 0 \\
\bottomrule
\end{tabular*}
\caption{\benchmark dataset audit. Counts are taken from the validation report and expected-document audit.}
\label{tab:appendix-release-audit}
\end{table}

\begin{table}[!tbp]
\centering
\scriptsize
\setlength{\tabcolsep}{2.4pt}
\begin{tabular*}{\columnwidth}{@{\extracolsep{\fill}}lccccc@{}}
\toprule
Split & QA & Dom. & Evid. & Zero & Oracle \\
\midrule
Full dataset & 3,000 & $10{\times}300$ & 3,000/3,000 & \maingemmazero/\maingeminizero & \maingemmaoracle/\maingeminioracle \\
Codex & 1,500 & $10{\times}150$ & 1,500/1,500 & \codexgemmazero/\codexgeminizero & \codexgemmaoracle/\codexgeminioracle \\
Claude & 1,500 & $10{\times}150$ & 1,500/1,500 & \claudegemmazero/\claudegeminizero & \claudegemmaoracle/\claudegeminioracle \\
\bottomrule
\end{tabular*}
\caption{Supplementary source-level calibration. Zero columns report \texttt{gemma2:9b}/Gemini~2.5 Flash zero-shot solvability without supporting chat evidence; oracle columns give the model the relevant context.}
\label{tab:appendix-split-calibration}
\end{table}

Each accepted record passes the same construction and filtering loop used in the main paper: scenario seed generation, chat and multiple-choice realization, deterministic normalization, adversarial refinement, exact evidence alignment, oracle answerability, and zero-shot hardness checks. The construction-log aggregates report \machineaccept\% acceptance after deterministic machine filtering and \humanaccept\% acceptance after review by \humanreviewers{} human reviewers. The validation report confirms the final dataset properties rather than these intermediate construction-log rates: answer labels are globally balanced and nearly balanced within each topic, with each domain containing 300 examples and each split contributing 150 examples per domain.

\paragraph{Construction and validation details.}
Machine filters check schema validity, option parseability, answer-label consistency, exact evidence alignment, oracle answerability, and zero-shot hardness. Three human reviewers then assess answerability, uniqueness, evidence grounding, and distractor ambiguity. Per-stage acceptance rates and final dataset counts are reported in the tables above.

\paragraph{Multiple-choice evaluation protocol.}
\label{app:mcq-protocol}
\benchmark uses deterministic four-way MCQ to make evidence-grounded grading auditable. This design isolates memory recovery from generation-style variation, which otherwise conflates surface style, answer length, and evaluator behavior with memory ability. In each item, distractors are plausible under generic priors but invalidated by local evidence, so a fluent answer that misses the operative constraint should fall into the plausible-distractor trap. Conversely, a system that retrieves the correct evidence span but produces awkward prose should still receive credit for selecting the correct option. Open-ended assistant-memory evaluation remains complementary; \benchmark focuses on reproducible episode-integrity diagnosis rather than long-form explanation quality.

\subsection{Blind Human Realism Audit}

The final human audit uses 300 stratified examples and three anonymous annotators. All three annotators completed all examples, producing 900 valid annotation rows. Each auditor saw the question, four answer options, and the complete item-level source dialogue, but not the 128k noise-packed runtime context; gold answers, expected evidence, reasoning, and system outputs were hidden. The displayed dialogues had a median of 7 turns and approximately 136 estimated tokens. The audit therefore assesses item-level realism, answerability, ambiguity risk, constraint plausibility, and human recoverability from the source dialogue, rather than human retrieval difficulty over noisy long contexts. Table~\ref{tab:appendix-human-realism} reports item-level means on a 1--5 Likert scale; lower is better for ambiguity risk. The majority answer agrees with gold in 296 of 299 majority-valid examples (\(99.0\%\)); the remaining audit set contains one no-majority case and three wrong-majority cases. Annotator-level response balance, pairwise agreement, stress-label distribution, and per-topic realism are reported in Tables~\ref{tab:appendix-human-qc}, \ref{tab:appendix-human-pairwise}, \ref{tab:appendix-stress-distribution}, and \ref{tab:appendix-realism-topic}.

The single no-majority case is a Game-Novel item (\texttt{S300-204}) where annotators split across three options under overlapping long-range-bridge and constraint-trap cues. The three wrong-majority cases are Network-Hardware (\texttt{S300-084}), Engineering (\texttt{S300-213}), and Social-Personal (\texttt{S300-247}) items; all involve overlapping stress cues, and two have all three cue labels active. These cases are retained in the audit accounting rather than removed.

\begin{table}[!tbp]
\centering
\scriptsize
\setlength{\tabcolsep}{4.0pt}
\begin{tabular*}{\columnwidth}{@{\extracolsep{\fill}}lrrrr@{}}
\toprule
Metric & Mean & Median & Std. & Direction \\
\midrule
Naturalness & 3.80 & 3.67 & 0.25 & higher better \\
Answerability & 4.91 & 5.00 & 0.17 & higher better \\
Ambiguity risk & 1.45 & 1.33 & 0.32 & lower better \\
Constraint plausibility & 3.99 & 4.00 & 0.29 & higher better \\
\bottomrule
\end{tabular*}
\caption{Overall blind human realism audit on 300 stratified examples with three anonymous annotators.}
\label{tab:appendix-human-realism}
\end{table}

\begin{table}[!tbp]
\centering
\scriptsize
\setlength{\tabcolsep}{3.6pt}
\resizebox{\columnwidth}{!}{%
\begin{tabular}{lrrrrrr}
\toprule
Ann. & Gold Acc. & Max ans. & Natural & Answerable & Ambig. & Plausible \\
\midrule
Ann. 1 & 99.0 & 25.7 & 4.17 & 4.90 & 1.19 & 3.98 \\
Ann. 2 & 91.0 & 30.3 & 4.04 & 4.87 & 1.92 & 4.08 \\
Ann. 3 & 96.3 & 25.3 & 3.20 & 4.95 & 1.25 & 3.90 \\
\bottomrule
\end{tabular}
}
\caption{Annotator-level descriptive checks for the three-annotator human audit. Gold Acc. is agreement with the benchmark answer; Max ans. is the largest selected-answer share, used as a response-balance check.}
\label{tab:appendix-human-qc}
\end{table}

\begin{table}[!tbp]
\centering
\scriptsize
\setlength{\tabcolsep}{3.2pt}
\resizebox{\columnwidth}{!}{%
\begin{tabular}{lrrrrrrr}
\toprule
Pair & Ans. agr. & Ans. \(\kappa\) & Primary agr. & State & Long & Trap & Cons. gold \\
\midrule
Ann. 1--2 & 91.3 & 0.884 & 56.7 & 43.3 & 83.0 & 84.7 & 272/274 \\
Ann. 1--3 & 96.0 & 0.947 & 16.3 & 45.3 & 83.0 & 100.0 & 287/288 \\
Ann. 2--3 & 89.0 & 0.853 & 32.7 & 57.3 & 72.0 & 84.7 & 265/267 \\
\bottomrule
\end{tabular}
}
\caption{Pairwise agreement in the final human audit. Answer-choice agreement is high, including chance-corrected pairwise Cohen's \(\kappa\); primary-stress agreement is lower because many examples contain overlapping stress cues. Cons. gold reports gold agreement among pairwise majority-resolved cases.}
\label{tab:appendix-human-pairwise}
\end{table}

\begin{table}[!tbp]
\centering
\scriptsize
\setlength{\tabcolsep}{4.0pt}
\begin{tabular*}{\columnwidth}{@{\extracolsep{\fill}}llrr@{}}
\toprule
View & Label & Count & Rate \\
\midrule
Multi-label cue & State overwrite & 196 & 65.3 \\
Multi-label cue & Long-range bridge & 291 & 97.0 \\
Multi-label cue & Constraint trap & 300 & 100.0 \\
Multi-label cue & Any multi-label & 298 & 99.3 \\
Multi-label cue & All three cues & 189 & 63.0 \\
\midrule
Primary cue & Constraint trap & 158 & 52.7 \\
Primary cue & State overwrite & 80 & 26.7 \\
Primary cue & Long-range bridge & 9 & 3.0 \\
Primary cue & Needs adjudication & 53 & 17.7 \\
\bottomrule
\end{tabular*}
\caption{Human stress-label distribution on the same 300 examples. Stress cues are intentionally multi-label; primary labels summarize the dominant cue only. Needs adjudication indicates cases where annotators did not form a stable dominant-stress label, not invalid examples.}
\label{tab:appendix-stress-distribution}
\end{table}

\begin{table}[!tbp]
\centering
\scriptsize
\setlength{\tabcolsep}{3.0pt}
\begin{tabular*}{\columnwidth}{@{\extracolsep{\fill}}lrrrr@{}}
\toprule
Topic & Natural & Answerable & Ambig. & Plausible \\
\midrule
Biomed & 3.66 & 4.86 & 1.49 & 4.00 \\
Biz-Ops & 3.79 & 4.92 & 1.31 & 4.13 \\
CS-Software & 3.74 & 4.94 & 1.34 & 4.09 \\
Daily-Life & 4.09 & 4.91 & 1.38 & 4.04 \\
Engineering & 3.72 & 4.90 & 1.42 & 4.13 \\
Finance & 3.79 & 4.88 & 1.41 & 4.06 \\
Game-Novel & 3.71 & 4.93 & 1.61 & 3.63 \\
Legal & 3.67 & 4.92 & 1.38 & 3.94 \\
Network & 3.72 & 4.90 & 1.33 & 3.94 \\
Social & 4.13 & 4.90 & 1.86 & 3.88 \\
\bottomrule
\end{tabular*}
\caption{Per-topic realism means in the 300-example human audit. Ambig. denotes ambiguity risk, where lower is better.}
\label{tab:appendix-realism-topic}
\end{table}

\section{Length-Controlled Runtime Evaluation Protocol}
\label{app:runtime}

\benchmark evaluates systems under a user-specified target constructed context length. This value is a benchmark-side packing target, not a model-window budget, a benchmark-internal upper bound, or a claim that every provider tokenizer assigns exactly the same number of native tokens. For each experiment, the builder creates a runtime package with the same selected records, noise, seeds, batch mapping, and executable files for every compared method. We therefore use the benchmark-side estimate for deterministic packing and use stored prompt/context text for any backend-specific tokenizer audit. The public repository provides an MIT-licensed UltraChat-derived default pool for direct use \citep{ding-etal-2023-ultrachat}; exact reproduction of the reported packages uses the pinned WildChat rebuild path and verifies the original noise hashes. Table~\ref{tab:appendix-runtime-protocol} summarizes the length-controlled packing protocol.

\begin{table}[!tbp]
\centering
\scriptsize
\setlength{\tabcolsep}{3.0pt}
\begin{tabular*}{\columnwidth}{@{\extracolsep{\fill}}p{0.30\columnwidth}p{0.64\columnwidth}@{}}
\toprule
Protocol item & Definition \\
\midrule
Token accounting & Benchmark-side estimate \(\mathrm{tokens}\approx1.3\times\) whitespace word count, used for deterministic packing and constructed-length reporting, not for asserting exact provider-native token parity. \\
Length scaling & No built-in length cap; practical limits are distractor-pool size and computational budget. This paper reports settings through 1M to match evaluated native-context backends. \\
Full-corpus regime & Used when selected truth tokens fit inside the target context length; the full selected truth background is retained and noise fills remaining space. \\
Length-batched regime & Used when selected truth tokens exceed the target length; records are deterministically partitioned into budget-controlled batches. \\
Packing rule & Deterministic capacity-constrained best fit: records are ordered by descending truth-token count and then by question name. \\
Truth cap & In length-batched mode, the default truth-cap ratio is \(0.82\), leaving room for noise and prompt/interface overhead. \\
Noise fill & Noise blocks are deterministically shuffled with the noise seed and inserted identically for all methods evaluated on the package. \\
Runtime outputs & Each package contains a manifest, stats, selected IDs, \texttt{GROUND\_TRUTH\_HISTORY}, \texttt{EVALUATION\_QUERIES}, and \texttt{NOISE}. \\
\bottomrule
\end{tabular*}
\caption{Length-controlled evaluation protocol used by the deterministic runtime-package builder.}
\label{tab:appendix-runtime-protocol}
\end{table}

For the full \targetsize{}-question dataset, the truth corpus is approximately \fulltruthtokens{} benchmark-side tokens. Thus 128k and 256k experiments naturally instantiate length-batched evaluation, whereas 512k and larger targets can enter the full-corpus regime. The main 128k comparison uses four deterministic batches with the same mix seed and truth-cap policy for every retrieval system. Beyond the reported settings, longer packages can be generated by drawing additional distractor turns; future longer-window models can be evaluated without changing the benchmark logic.

\section{Method and Baseline Implementation Details}
\label{app:method-details}

All methods are evaluated under the same persistent writeback setting. None of the retrieval baselines receives dataset-provided gold blocks, future turns, or method-specific noise. The key implementation differences are summarized in Table~\ref{tab:appendix-method-details}.

\begin{table*}[!tbp]
\centering
\scriptsize
\setlength{\tabcolsep}{3.5pt}
\begin{tabular*}{\textwidth}{@{\extracolsep{\fill}}p{0.16\textwidth}p{0.22\textwidth}p{0.27\textwidth}p{0.25\textwidth}@{}}
\toprule
System & Retrieval unit & Context assembly & Purpose in comparison \\
\midrule
Standard RAG top-5 & Individual retrieved chunks/turns & Top five retrieved units are passed to the answer backend. & Tests whether short semantic retrieval alone can recover the decisive evidence. \\
\hybridrag & Dense chunks + BM25 sparse hits + neighboring turns & Dense and sparse hits are fused with reciprocal-rank fusion, expanded with local neighbors, and packed under a matched context cap. & Tests whether lightweight hybrid chunk retrieval closes the episode-reconstruction gap. \\
\tunedrag & Dense/sparse candidates + HyDE + cross-encoder reranking & Adds query expansion, RRF fusion, BGE reranking, parent-window expansion, and MMR packing under the same GPT-4o-mini 128k task. & Tests whether a strongly tuned but non-episodic chunk pipeline can close the gap, and exposes its reranking cost. \\
RAPTOR strict no-block & Hierarchical summaries built without gold blocks & Retrieved hierarchy outputs are mapped into the same no-block evaluation regime. & Tests whether hierarchical abstraction alone solves interleaved conversational memory. \\
\textsc{MemGPT} paper-default & Explicit memory-management substrate & Retrieved memory material is inserted under the same answer and writeback protocol. & Tests whether higher recall from a memory-style substrate converts into final accuracy. \\
\textsc{HippoRAG} paper-default & Graph-style retrieval substrate & Retrieved graph/memory evidence is evaluated with the same question set and scoring protocol. & Tests transfer of graph-centric memory retrieval to flat-thread writeback evaluation. \\
Official \tsim & Raw turns, reconstructed episodes, and L2 semantic clusters & Raw and summary hits contribute to episode scores; L2 routes to related episodes; final top episodes form the prompt. & Tests block-free episode reconstruction and multi-granularity memory organization. \\
\bottomrule
\end{tabular*}
\caption{High-level implementation distinctions for the main retrieval and memory systems.}
\label{tab:appendix-method-details}
\end{table*}

\paragraph{Tuned Hybrid-Rerank control.}
The \tunedrag control combines BM25 sparse retrieval \citep{robertson-zaragoza-2009-bm25}, BGE dense retrieval and reranking \citep{xiao-etal-2023-bge}, HyDE query rewriting \citep{gao-etal-2023-hyde}, reciprocal-rank fusion \citep{cormack-etal-2009-rrf}, parent-window expansion, and MMR packing \citep{carbonell-goldstein-1998-mmr}. It is non-episodic: it does not use semantic episode segmentation or multi-granularity \tsim memory.

All dense \tsim memory levels use \texttt{BAAI/bge-large-en-v1.5} through SentenceTransformers. Each reconstructed episode \(e=[s_e,t_e]\) is represented by its original per-turn embeddings and a deterministic summary text that prefixes a relative episode-recency tag and episode id to episode text truncated to 1,200 characters. Cluster summaries deterministically combine recent member-episode summary texts. Raw turns, episode summaries, and cluster-summary text embeddings are stored in separate Chroma HNSW indices with cosine distance; centroid vectors remain in memory only for episode-to-cluster assignment and merging. No LLM calls are used to construct the summary or cluster views. The implementation keeps \texttt{all-MiniLM-L6-v2} only for the retrieval sensitivity diagnostic reported in Appendix~\ref{app:additional-results}, not for the main answer-model runs. Table~\ref{tab:drift-algorithm} gives compact pseudocode for the streaming semantic-drift segmenter.

\paragraph{M1 formal definitions.}
Let \(x_i\in\mathbb{R}^d\) be the normalized embedding of turn \(i\), \(R_i\) the recent turns still inside the current episode, and \(L_i\) the current episode length. The segmenter uses:
\begin{align*}
c_i &= \operatorname{norm}\!\left(\frac{1}{|R_i|}\sum_{j\in R_i}x_j\right),\\
s_i &= \cos(x_i,c_i)\\
&\quad +b\,\mathbb{I}[z_{i-1}=\mathrm{model},z_i=\mathrm{user}],\\
\operatorname{cut}(i) &=
\begin{aligned}[t]
&\mathbb{I}\!\left[s_i<\theta_s \land L_i\ge L_{\min}\right]\\
&{}\lor \mathbb{I}\!\left[L_i\ge L_{\max}\right].
\end{aligned}
\end{align*}
Here \(z_i\) denotes the speaker, \(\theta_s=0.70\), and \(b=0.03\). The first guard opens a boundary only after the current episode reaches the minimum length, while the second prevents oversized episodes.

\begin{table}[!tbp]
\centering
\scriptsize
\begin{tabular*}{\columnwidth}{@{}p{\columnwidth}@{}}
\toprule
\textbf{Streaming semantic-drift segmentation in Official \tsim} \\
\midrule
1. Flatten the mixed conversation into a turn stream and encode each turn as a normalized dense vector. \\
2. Maintain the current segment start \(s\) and token count \(L\). \\
3. For incoming turn \(i\), average only the recent turns still inside the current segment to form local center \(c_i\). \\
4. Score the newest turn by cosine similarity to \(c_i\), plus a small bonus for a \texttt{model} \(\rightarrow\) \texttt{user} transition. \\
5. If \(L \ge \tau_{\min}\) and the score falls below similarity threshold \(\theta_s\), close the segment before turn \(i\). \\
6. If \(L \ge \tau_{\max}\), force a boundary even if semantic drift is weak. \\
7. Continue streaming without revisiting earlier boundaries. \\
\bottomrule
\end{tabular*}
\caption{Compact pseudocode view of the semantic-drift segmenter used by Official \tsim. The final system uses a simple local-window cosine rule rather than a heavier offline clustering procedure for boundary decisions.}
\label{tab:drift-algorithm}
\end{table}

The four terms in the main-text scoring equation are:
\begin{align}
R_r(e,q) &= \sum_{r \in H_r(e,q)} \mathrm{sim}(q,r), \\
R_s(e,q) &= \sum_{s \in H_s(e,q)} \mathrm{sim}(q,s), \\
R_l(e,q) &= \sum_{c \in H_l(q)}
\mathbb{I}\!\left[e \in \operatorname{Expand}(c)\right] \nonumber\\
&\quad \cdot \lambda_l^{\operatorname{rank}(e;c)}
\operatorname{sim}(q,c), \\
R_{\mathrm{sem}}(e,q)
&= \sum_{e' \in \operatorname{Seeds}(q)}
\lambda_{\mathrm{sem}}\,\operatorname{sim}(q,e') \nonumber\\
&\quad \cdot \operatorname{sim}(e',e).
\end{align}
Here \(H_r(e,q)\) and \(H_s(e,q)\) are raw-turn and summary hits attached to episode \(e\), while \(H_l(q)\) is the set of retrieved L2 clusters. \(\operatorname{Expand}(c)\) returns the small set of episodes attached to cluster \(c\), and \(\operatorname{Seeds}(q)\) are the top local episode candidates used for semantic expansion.

\section{\tsim Configuration Selection and Ablation}
\label{app:selection-ablation}

All reported SCALE-QA \tsim results use one frozen configuration across answer backends rather than backend-specific retuning. Because \tsim is an architecture rather than an end-to-end trained model, its scalar coefficients are calibrated constants rather than learned parameters. The selection protocol emphasizes evidence-reconstruction stability over small-sample answer accuracy alone. Table~\ref{tab:appendix-selection-protocol} summarizes the frozen-configuration selection stages, and Table~\ref{tab:appendix-tsim-constants} lists the exact constants used in the reported SCALE-QA runs.

\begin{table}[!tbp]
\centering
\scriptsize
\setlength{\tabcolsep}{3.0pt}
\begin{tabular*}{\columnwidth}{@{\extracolsep{\fill}}p{0.26\columnwidth}p{0.18\columnwidth}p{0.48\columnwidth}@{}}
\toprule
Stage & Data & Purpose \\
\midrule
Search & 300 QA & Fast screening of candidate retrieval configurations. \\
Confirm & 600 QA & Check stability across 64k, 128k, 256k, and 512k targets. \\
Answer validation & 200 QA & Confirm that retrieval gains transfer to answer accuracy. \\
Full retrieval check & 3,000 QA & Verify evidence coverage and context stability at 128k. \\
\bottomrule
\end{tabular*}
\caption{Frozen-configuration selection protocol for the SCALE-QA experiments.}
\label{tab:appendix-selection-protocol}
\end{table}

Table~\ref{tab:parameter-stability} summarizes representative neighborhood stability from the bounded development sweep. The table is not meant to present every tried configuration or to claim that the reported constants are uniquely optimal. Instead, it shows that several nearby settings around the same semantic-drift threshold, retrieval breadth, and L2 routing weight remain strong on disjoint search, confirmation, and answer-validation splits. The reported setting is frozen because it gives the best balance of evidence coverage, answer transfer, and compact context; search-only peaks are not selected unless they also confirm under the longer-context stability check.

\begin{table*}[!tbp]
\centering
\scriptsize
\setlength{\tabcolsep}{2.8pt}
\begin{tabular*}{\textwidth}{@{\extracolsep{\fill}}p{0.23\textwidth}p{0.25\textwidth}ccccc@{}}
\toprule
Candidate & Representative variation & Search Hit & Confirm Hit & Answer Hit & Answer Acc & Answer Ctx \\
\midrule
Official calibrated & \(\theta=.70,k_f=5,k_r/k_s=28/20,w_l=.75\) & 86.33 & \textbf{79.33} & \textbf{87.50} & 78.50 & 1011.6 \\
High-threshold compact & \(\theta=.74,k_r=44,k_s=16,w_l=.85\) & 85.67 & 77.62 & 86.00 & \textbf{79.00} & 757.7 \\
Lean memory stack & \(k_f=4,k_r/k_s=36/12,w_l=.65\) & 84.00 & 76.58 & \textbf{87.50} & 77.50 & 818.2 \\
Lean L2-exp \(=1\) & Same as lean, one L2-to-episode expansion & 83.67 & 76.85 & \textbf{87.50} & 77.50 & 818.1 \\
\bottomrule
\end{tabular*}
\caption{Representative neighborhood stability from the development sweep. Search300 and confirm600 report CL Hit; answer200 reports CL Hit, Accuracy, and average retrieved context tokens. The answer-validation prompt constrains outputs to valid answer choices, so the reported answer accuracy follows the paper's main Accuracy convention.}
\label{tab:parameter-stability}
\end{table*}

\begin{figure}[!tbp]
\centering
\includegraphics[width=\columnwidth]{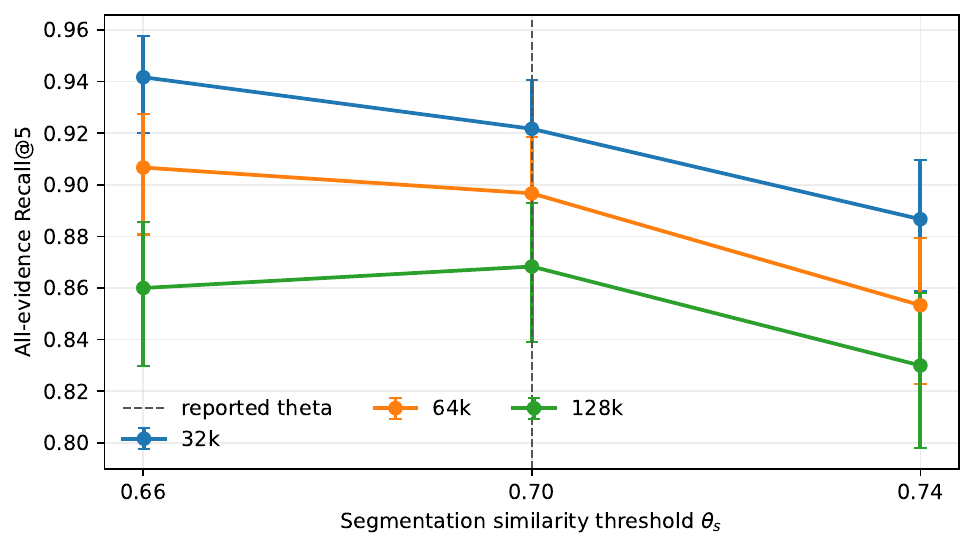}
\caption{Single-variable M1 threshold sensitivity on the frozen confirmation split. Points report all-evidence Recall@5 and bars show Wilson 95\% confidence intervals; the question IDs, runtime construction, and all non-threshold settings are fixed.}
\label{fig:theta-sensitivity}
\end{figure}

\paragraph{M1 threshold sensitivity.}
Figure~\ref{fig:theta-sensitivity} isolates the segmentation threshold on the same 600 confirmation questions at 32k, 64k, and 128k, with all other reported settings fixed. The paired-bootstrap 95\% intervals for the Recall@5 difference between \(\theta_s=.66\) and the reported \(.70\) are \([-0.002,0.042]\), \([-0.013,0.035]\), and \([-0.038,0.023]\), respectively. All include zero, showing stable retrieval near the reported threshold across context lengths.

\begin{table}[!tbp]
\centering
\scriptsize
\setlength{\tabcolsep}{3.0pt}
\begin{tabular*}{\columnwidth}{@{\extracolsep{\fill}}p{0.30\columnwidth}p{0.64\columnwidth}@{}}
\toprule
Component & Frozen reported constants \\
\midrule
Segmenter & \texttt{min\_tokens}=120, \texttt{max\_tokens}=320, \texttt{recent\_window}=4, \(\theta_s=0.70\), speaker bonus \(b=0.03\), minimum two-message buffer. \\
Retrieval breadth & \(k_r=28\), \(k_s=20\), \(k_l=2\), \(k_{\mathrm{final}}=5\). \\
Ranking weights & \(w_r=1.15\), \(w_s=1.20\), \(w_l=0.75\), \(\lambda_{\mathrm{sem}}=0.55\), \(\lambda_l=0.70\). \\
Expansion & Cluster expansion \(k=2\), L2-to-episode expansion \(k=2\), cluster threshold \(0.42\), soft margin \(0.08\), temporal expansion hops \(0\). \\
Prompt assembly & Reconstructed episodes capped at eight messages with two-message overlap; L2 cluster summaries route and boost candidates but are not inserted directly into the prompt. \\
\bottomrule
\end{tabular*}
\caption{Exact \tsim constants used in the reported SCALE-QA runs. These values are moved out of the main text to avoid visually overloading the method narrative with implementation constants.}
\label{tab:appendix-tsim-constants}
\end{table}

\section{Additional Experimental Results}
\label{app:additional-results}

Table~\ref{tab:cross-backend-domain} expands the aggregate cross-backend comparison from Table~\ref{tab:cross-backend-metrics} into all ten domains. It is an audit table rather than an additional leaderboard.

\begin{table}[!tbp]
\centering
\tiny
\setlength{\tabcolsep}{2.5pt}
\resizebox{\columnwidth}{!}{%
\begin{tabular}{lrrrrrr}
\toprule
\multicolumn{7}{l}{\textbf{Gemma2:9b}} \\
Domain & \tsim & Std. RAG & Hybrid & RAPTOR & MemGPT & HippoRAG \\
\midrule
CS & 79.0 & 29.3 & 33.7 & 48.0 & 66.7 & 25.3 \\
Net & 77.3 & 26.7 & 33.7 & 49.3 & 67.0 & 27.0 \\
Fin & 54.3 & 22.7 & 29.3 & 32.7 & 46.7 & 23.3 \\
Legal & 60.0 & 22.3 & 28.0 & 36.3 & 56.3 & 26.7 \\
Bio & 74.7 & 25.3 & 36.7 & 39.7 & 66.7 & 21.3 \\
Eng & 69.7 & 23.0 & 30.3 & 35.3 & 55.0 & 22.0 \\
Biz & 64.3 & 25.3 & 32.3 & 36.3 & 54.0 & 29.7 \\
Social & 68.0 & 20.7 & 25.3 & 36.0 & 56.3 & 18.7 \\
Game & 68.3 & 18.0 & 26.3 & 36.0 & 58.3 & 33.0 \\
Daily & 80.7 & 31.0 & 35.3 & 57.3 & 73.7 & 25.7 \\
All & \textbf{69.6} & 24.4 & 31.1 & 40.6 & 60.1 & 25.2 \\
\midrule
\multicolumn{7}{l}{\textbf{Gemini 2.5 Flash}} \\
CS & 85.3 & 37.7 & 35.7 & 66.0 & 91.7 & 34.7 \\
Net & 93.0 & 30.7 & 33.3 & 61.3 & 87.0 & 34.3 \\
Fin & 67.0 & 26.3 & 28.0 & 37.7 & 58.3 & 31.3 \\
Legal & 79.0 & 27.7 & 30.3 & 45.7 & 67.3 & 38.0 \\
Bio & 86.3 & 34.3 & 43.7 & 62.3 & 84.0 & 35.0 \\
Eng & 79.3 & 28.0 & 32.0 & 50.3 & 73.3 & 28.0 \\
Biz & 72.0 & 28.0 & 30.7 & 45.7 & 67.0 & 39.0 \\
Social & 75.0 & 25.7 & 28.7 & 43.3 & 64.0 & 29.0 \\
Game & 83.3 & 29.0 & 34.7 & 50.3 & 69.7 & 47.0 \\
Daily & 81.3 & 31.3 & 31.3 & 63.0 & 83.7 & 28.7 \\
All & \textbf{80.2} & 29.8 & 32.8 & 52.6 & 74.6 & 34.4 \\
\midrule
\multicolumn{7}{l}{\textbf{GPT-4o-mini}} \\
CS & 85.0 & 37.3 & 34.7 & 56.0 & 60.3 & 33.7 \\
Net & 80.3 & 27.7 & 31.7 & 50.3 & 59.7 & 29.3 \\
Fin & 66.3 & 25.0 & 27.7 & 35.3 & 39.7 & 31.3 \\
Legal & 72.3 & 22.3 & 29.3 & 39.0 & 45.7 & 35.7 \\
Bio & 77.0 & 29.0 & 37.7 & 43.3 & 54.7 & 31.0 \\
Eng & 72.3 & 26.7 & 30.7 & 41.3 & 46.3 & 26.0 \\
Biz & 70.3 & 23.7 & 29.7 & 38.0 & 46.0 & 36.0 \\
Social & 66.3 & 22.0 & 25.0 & 33.3 & 46.0 & 25.7 \\
Game & 64.0 & 25.3 & 34.0 & 37.0 & 40.3 & 34.0 \\
Daily & 83.7 & 32.3 & 30.7 & 56.7 & 63.7 & 28.3 \\
All & \textbf{73.8} & 27.1 & 31.1 & 43.0 & 50.2 & 31.0 \\
\bottomrule
\end{tabular}
}
\caption{Supplementary per-domain Accuracy for the 128k cross-backend comparison. Hybrid denotes \hybridrag. Domain abbreviations: CS = software, Net = network/hardware, Fin = finance, Bio = biomedicine, Eng = engineering, Biz = business operations.}
\label{tab:cross-backend-domain}
\end{table}

\begin{table*}[!tbp]
\centering
\scriptsize
\setlength{\tabcolsep}{3.2pt}
\resizebox{\textwidth}{!}{%
\begin{tabular}{lrrrrrrr}
\toprule
Stress view & \(N\) & Std. RAG & Hybrid-RRF & RAPTOR & MemGPT & HippoRAG & \tsim \\
\midrule
All audited & 300 & 26.7/5.7 & 29.0/9.3 & 41.7/29.7 & 52.7/58.3 & 29.3/8.7 & \textbf{73.7/78.7} \\
State overwrite & 196 & 24.0/3.6 & 27.0/8.2 & 38.3/26.5 & 49.5/56.6 & 28.6/7.1 & \textbf{70.9/76.0} \\
Long-range bridge & 291 & 26.8/5.8 & 28.9/8.9 & 42.3/29.9 & 52.9/58.1 & 29.6/8.6 & \textbf{73.9/78.4} \\
Constraint trap & 300 & 26.7/5.7 & 29.0/9.3 & 41.7/29.7 & 52.7/58.3 & 29.3/8.7 & \textbf{73.7/78.7} \\
\bottomrule
\end{tabular}
}
\caption{Full stress-type method audit on the 300-example three-annotator human-audited subset under GPT-4o-mini at 128k. Cells report Accuracy / CL Hit. Stress views are multi-label, so counts are not expected to sum to 300. Full Context is omitted because only 11 audited items have matched rows in this diagnostic slice.}
\label{tab:appendix-stress-methods}
\end{table*}

\subsection{Episode Reconstruction Diagnostics}

We evaluate episode reconstruction through exact evidence traces rather than subjective discourse-boundary labels. All-evidence recall@5 asks whether the union of the top five retrieved units covers all gold evidence and is computed over all 3,000 questions. Co-containment asks, for questions whose gold evidence spans multiple snippets, whether any one returned unit contains all decisive evidence. The columns therefore have different denominators. For \tsim, a unit is a reconstructed episode; for the controls, it is a retrieved chunk or fixed window.

\begin{table}[!tbp]
\centering
\scriptsize
\setlength{\tabcolsep}{3.0pt}
\begin{tabular*}{\columnwidth}{@{\extracolsep{\fill}}lrr@{}}
\toprule
Retrieved unit & All recall@5 & Co-contain \\
\midrule
\tsim episode & \textbf{0.810} & \textbf{0.890} \\
Fixed 128-token window & 0.719 & 0.675 \\
Fixed 256-token window & 0.647 & 0.826 \\
Fixed 320-token window & 0.577 & 0.831 \\
Standard RAG chunk & 0.456 & 0.004 \\
\bottomrule
\end{tabular*}
\caption{Exact-evidence reconstruction diagnostics on the full 3,000-question runtime. All recall@5 uses all questions; co-containment uses the multi-snippet subset. Larger fixed windows improve co-containment but reduce recall, whereas \tsim improves both.}
\label{tab:segmentation-diagnostic}
\end{table}

For each of the 3,000 questions, a raw- or summary-view hit is counted when at least one item returned by that view maps to a reconstructed episode containing a matched gold-evidence snippet; an L2 hit is counted when a retrieved cluster contains such an episode. Using the frozen SCALE-QA retrieval depths (28 raw, 20 summary, and 2 L2), the corresponding evidence-episode hit rates are \(0.638\), \(0.861\), and \(0.708\), respectively. The episode-level view most often recovers evidence missed at raw-turn granularity, while cluster retrieval supplies complementary routing evidence.

\begin{table}[!tbp]
\centering
\scriptsize
\setlength{\tabcolsep}{3.5pt}
\begin{tabular*}{\columnwidth}{@{\extracolsep{\fill}}llr@{}}
\toprule
Method & Embedder & All recall@5 \\
\midrule
Standard RAG & BGE-large & 0.456 \\
Standard RAG & MiniLM & 0.408 \\
\tsim & BGE-large & \textbf{0.810} \\
\tsim & MiniLM & 0.649 \\
\bottomrule
\end{tabular*}
\caption{Retrieval-only embedding sensitivity on the same 3,000 questions. MiniLM denotes \texttt{all-MiniLM-L6-v2}; BGE-large denotes \texttt{BAAI/bge-large-en-v1.5}.}
\label{tab:embedding-sensitivity}
\end{table}

\subsection{LongMemEval-S Transfer Diagnostic}

We evaluate one LongMemEval-specific adaptation uniformly across all 500 LongMemEval-S cleaned V1 questions, without per-question or question-type routing. The configuration was selected using full-set retrieval criteria, so we report this as a transductive transfer diagnostic rather than a held-out generalization estimate. All methods use the same question IDs, Gemini~2.5 Flash at temperature zero, and the official GPT-4o judged-response protocol. At inference time, gold answers, evidence annotations, and question types are hidden from retrieval and answering. \tsim and the two no-boundary controls receive the same chronologically ordered flat turn stream; only the session-level diagnostic uses official session boundaries. The fixed-chunk control partitions this stream into non-overlapping chunks of consecutive complete turns with a 1,100-token target, then retrieves the top five with BM25.

\paragraph{Adaptation configuration.} Relative to Table~\ref{tab:appendix-tsim-constants}, the segmenter uses \(\theta=.74\), 160--480-token episodes, a two-episode recent window, and speaker bonus \(0.06\); retrieval uses \(k_r/k_s/k_l/k_{\mathrm{final}}=16/24/1/8\), weights \(w_r/w_s/w_l=1.5/1.5/.35\), and temporal decay \(0.5\). Other active settings follow Table~\ref{tab:appendix-tsim-constants}; each question starts with fresh memory, QA writeback is disabled, and L2 context is not inserted into the answer prompt.

\begin{table*}[!tbp]
\centering
\small
\setlength{\tabcolsep}{4.0pt}
\begin{tabular*}{\textwidth}{@{\extracolsep{\fill}}p{0.25\textwidth}p{0.27\textwidth}rrr@{}}
\toprule
Method & Boundary information & \shortstack{Judged\\Acc. (\%)} & \shortstack{All-evidence\\recall (\%)} & \shortstack{Avg. answer-\\context tokens} \\
\midrule
\tsim & None; episodes reconstructed & \textbf{71.0} & \textbf{84.04} & 4,723.5 \\
BM25 fixed-chunk top-5 & None & 61.2 & 65.74 & 4,616.3 \\
BGE turn top-5 & None & 56.6 & 70.64 & 340.8 \\
BM25 session top-5 & Official session boundaries & 74.4 & 76.60 & 13,086.5 \\
\bottomrule
\end{tabular*}
\caption{Targeted transductive diagnostic on LongMemEval-S cleaned V1 (500 questions). Judged accuracy follows the official free-form response protocol; all-evidence recall is computed on the 470 non-abstention questions with annotated evidence, and average context denotes answer-context tokens. The session row is boundary-assisted and not information-condition matched.}
\label{tab:longmemeval-transfer}
\end{table*}

Against the context-matched fixed-chunk control, \tsim improves accuracy by \(9.8\) points (paired-bootstrap 95\% CI: \([+5.2,+14.4]\)) and all-evidence recall by \(18.30\) points. Against BGE turn retrieval, the accuracy gain is \(14.4\) points (95\% CI: \([+10.2,+18.6]\)). The boundary-assisted session diagnostic uses \(2.77\times\) more answer context; its \(3.4\)-point accuracy advantage over \tsim is not significant in this evaluation (95\% CI for \tsim minus session: \([-7.2,+0.6]\), McNemar \(p=0.1109\)). Together, the boundary-free comparisons isolate transfer of episode reconstruction, while the session row remains a boundary-assisted diagnostic.
\section{Statistical and Token Accounting Audit}
\label{app:statistical-audit}
\subsection{System Cost and Qualitative Diagnostics}

Table~\ref{tab:system-cost} reports retrieval-side accounting on the full 3,000-question 128k runtime. The storage values are lower bounds for 32-bit vectors and exclude HNSW metadata and document-store overhead. The deterministic summary and cluster views require no LLM calls. Answer-context usage is reported under the main experimental protocol in Table~\ref{tab:cross-backend-metrics}; we keep it separate because the cost replay uses a different token-accounting path.

\begin{table*}[!tbp]
\centering
\scriptsize
\setlength{\tabcolsep}{3.6pt}
\resizebox{\textwidth}{!}{%
\begin{tabular}{lrrrrrrrr}
\toprule
System & Views & Vectors & Vector MB & Text amp. & Ingest ms/turn & Ret. p50 ms & Ret. p95 ms & Summary API \\
\midrule
Standard RAG & 1 & 5,554 & 21.7 & 1.056\(\times\) & 4.178 & 21.478 & 36.791 & 0 \\
\tsim & 3 & 6,281 & 24.5 & 2.087\(\times\) & 24.986 & 141.093 & 159.978 & 0 \\
\bottomrule
\end{tabular}
}
\caption{Measured retrieval-side system accounting. Vector MB is a lower-bound dense-vector estimate; text amplification counts indexed textual views relative to the source history. Latencies characterize this implementation and hardware, not a hardware-normalized leaderboard.}
\label{tab:system-cost}
\end{table*}

Under the evaluated protocol, \tsim pays the higher embedding, indexing, and retrieval cost shown above while reducing answer-time context to about \(1\)k tokens in the GPT-4o-mini main comparison, versus \(3.0\)k for the tuned non-episodic control (Table~\ref{tab:cross-backend-metrics}). Provider-side KV-cache reuse could change Full Context economics in deployment, so this is protocol accounting rather than a universal crossover claim.

\begin{table*}[!tbp]
\centering
\scriptsize
\setlength{\tabcolsep}{4.0pt}
\begin{tabular*}{\textwidth}{@{\extracolsep{\fill}}p{0.20\textwidth}p{0.23\textwidth}p{0.49\textwidth}@{}}
\toprule
Failure type & Example & Diagnostic observation \\
\midrule
Episode-incomplete retrieval & \path{codex:Network-Hardware-075} & Standard RAG retrieves related text but omits the neighboring license clause that makes the local decision operative. \\
Fixed-window low SNR & \path{claude-code:Network-Hardware-032} & A fixed window contains the evidence within 1,755 tokens but at lower evidence density than the 1,127-token reconstructed episode. \\
Answer-side override misuse & \path{claude-code:Biomed-003} & \tsim retrieves the decisive episode, yet the answer model follows a plausible public default instead of the local override. \\
Routing/packing miss & \path{codex:Biomed-124} & Two gold units are required; final selected context covers only one, exposing a residual routing and packing failure. \\
\bottomrule
\end{tabular*}
\caption{Compact qualitative cases illustrating retrieval fragmentation, low signal-to-noise windows, answer-side evidence misuse, and residual \tsim routing errors.}
\label{tab:qualitative-errors}
\end{table*}

The main text reports compact aggregate scores, but the raw result files are example-aligned. We therefore audit the final main-table comparisons with paired bootstrap tests over the same 3,000 questions. The \hybridrag runs passed validation with 3,000 rows, zero run errors, and no duplicate IDs; the separate \tunedrag control also passed validation with 3,000 rows and zero duplicate IDs. The backend-output split, paired bootstrap comparisons, and variance/token audit are reported in Tables~\ref{tab:backend-output-audit}, \ref{tab:paired-significance}, and \ref{tab:variance-token-audit}. Tables~\ref{tab:fullcontext-sanity} and~\ref{tab:prompt-robust-fullcontext} provide Full Context evidence-containment and prompt-template diagnostics.

\paragraph{Metric and backend notes.}
Accuracy is the fraction of examples for which the selected option matches the gold label under forced-choice four-way MCQ grading. The three main answer backends provide complementary checks: \answermodel{} for local reproducibility, Gemini~2.5 Flash for long-context commercial evaluation, and GPT-4o-mini for high-throughput closed-loop comparison and ablation.

\begin{table}[!tbp]
\centering
\scriptsize
\setlength{\tabcolsep}{3.0pt}
\resizebox{\columnwidth}{!}{%
\begin{tabular}{lrrrrr}
\toprule
Budget & Hit & Acc & HitRank & Lat. & CtxTok \\
\midrule
16k & 100.00 & 62.5 & 335.4 & 4.32 & 29,938 \\
32k & 100.00 & 57.8 & 669.2 & 4.57 & 60,145 \\
64k & 100.00 & 41.7 & 836.3 & 10.03 & 86,315 \\
128k & 100.00 & 29.8 & 711.7 & 14.61 & 88,425 \\
\bottomrule
\end{tabular}
}
\caption{Full Context sanity audit for Figure~\ref{fig:full-release-scaling}. Hit is evidence containment in the constructed prompt. The collapse is therefore not explained by missing gold evidence alone.}
\label{tab:fullcontext-sanity}
\end{table}

\paragraph{Prompt-robust Full Context diagnostic.}
Because Full Context can depend on prompt structure, we ran an independent no-writeback GPT-4o-mini diagnostic on a stratified subset. A 20-example development split selected an evidence-first prompt from vanilla, evidence-first, and question-first variants using a fixed accuracy-maximization rule; the selected prompt was then applied unchanged to a disjoint 100-example evaluation split, with all three systems evaluated on the same 100 examples. All Full Context variants use the same visible-history window for each example, so differences reflect prompt structure rather than context selection. We focus on three structurally distinct prompt templates---vanilla baseline, evidence-first instruction, and question-first instruction---rather than exhaustively varying decoding strategies, because the diagnostic targets whether prompt structure alone can recover Full Context performance.

\begin{table}[!tbp]
\centering
\scriptsize
\setlength{\tabcolsep}{3.0pt}
\begin{tabular*}{\columnwidth}{@{\extracolsep{\fill}}llrrrr@{}}
\toprule
System & Prompt & Rows & Acc. & Prompt tok. & Lat. \\
\midrule
Full Context & Vanilla & 100 & 49.0 & 98,339 & 5.52 \\
Full Context & Dev-selected & 100 & 49.0 & 98,298 & 2.80 \\
\tsim & -- & 100 & 77.0 & 1,389 & 2.40 \\
\bottomrule
\end{tabular*}
\caption{Prompt-robust Full Context diagnostic under GPT-4o-mini in an independent no-writeback setting. The dev-selected prompt is evidence-first, chosen on a disjoint 20-example development split and frozen before evaluation. Rows are the same stratified 100-example evaluation subset with balanced domains, answer labels, and logical-context quartiles. Prompt tokens and latency are provider-reported averages; \tsim uses its standard memory pipeline rather than a Full Context prompt template. Numbers are not directly comparable to the main stateful results in Table~\ref{tab:cross-backend-metrics} because this diagnostic uses a 100-example subset under a no-writeback evaluation regime; it isolates prompt-template effects on Full Context rather than re-evaluating system rankings.}
\label{tab:prompt-robust-fullcontext}
\end{table}

\begin{table}[!tbp]
\centering
\tiny
\setlength{\tabcolsep}{2.0pt}
\resizebox{\columnwidth}{!}{%
\begin{tabular}{llrrrrr}
\toprule
Backend & Method & Acc & Hit-C & Hit-W & Miss-C & Miss-W \\
\midrule
Gemma2:9b & Standard RAG & 24.4 & 5.2 & 2.1 & 10.6 & 47.7 \\
 & \hybridrag & 31.1 & 8.4 & 2.1 & 9.5 & 27.2 \\
 & RAPTOR & 40.6 & 25.2 & 10.0 & 14.6 & 47.0 \\
 & \textsc{MemGPT} & 60.1 & 48.6 & 14.0 & 11.4 & 25.9 \\
 & \textsc{HippoRAG} & 25.2 & 8.7 & 3.7 & 15.6 & 68.4 \\
 & \textbf{\tsim} & 69.6 & 58.2 & 12.4 & 11.3 & 17.7 \\
\midrule
Gemini 2.5 Flash & Standard RAG & 29.8 & 6.1 & 0.2 & 0.5 & 0.3 \\
 & \hybridrag & 32.8 & 10.0 & 0.2 & 0.7 & 0.6 \\
 & RAPTOR & 52.6 & 32.1 & 1.5 & 6.2 & 3.2 \\
 & \textsc{MemGPT} & 74.6 & 61.5 & 0.5 & 5.2 & 1.0 \\
 & \textsc{HippoRAG} & 34.4 & 11.4 & 0.3 & 1.8 & 1.5 \\
 & \textbf{\tsim} & 80.2 & 66.5 & 1.2 & 8.4 & 2.8 \\
\midrule
GPT-4o-mini & Standard RAG & 27.1 & 5.0 & 1.4 & 5.3 & 21.0 \\
 & \hybridrag & 31.1 & 8.7 & 1.7 & 1.8 & 5.5 \\
 & RAPTOR & 43.0 & 26.0 & 8.9 & 14.6 & 40.7 \\
 & \textsc{MemGPT} & 50.2 & 36.4 & 9.4 & 1.5 & 3.3 \\
 & \textsc{HippoRAG} & 31.0 & 7.7 & 2.5 & 3.6 & 7.3 \\
 & \textbf{\tsim} & 73.8 & 63.5 & 10.4 & 8.4 & 10.2 \\
\bottomrule
\end{tabular}
}
\caption{Backend-output audit for the 128k main comparison. Hit-C/W and Miss-C/W split closed-loop evidence hit by correct and wrong answers, all in percentages.}
\label{tab:backend-output-audit}
\end{table}

\begin{table}[!tbp]
\centering
\tiny
\setlength{\tabcolsep}{2.0pt}
\resizebox{\columnwidth}{!}{%
\begin{tabular}{llcccc}
\toprule
Backend & Baseline & Base & \tsim & \(\Delta\) & 95\% CI \\
\midrule
\answermodel & Standard RAG & 24.4 & 69.6 & +45.21 & [43.48, 47.03] \\
 & \hybridrag & 31.1 & 69.6 & +38.52 & [36.72, 40.24] \\
 & RAPTOR & 40.6 & 69.6 & +28.98 & [27.13, 30.76] \\
 & \textsc{MemGPT} & 60.1 & 69.6 & +9.53 & [7.90, 11.22] \\
 & \textsc{HippoRAG} & 25.2 & 69.6 & +44.42 & [42.41, 46.36] \\
\midrule
Gemini 2.5 Flash & Standard RAG & 29.8 & 80.2 & +50.33 & [48.98, 51.63] \\
 & \hybridrag & 32.8 & 80.2 & +47.34 & [45.88, 48.73] \\
 & RAPTOR & 52.6 & 80.2 & +27.59 & [26.02, 29.19] \\
 & \textsc{MemGPT} & 74.6 & 80.2 & +5.58 & [4.15, 6.97] \\
 & \textsc{HippoRAG} & 34.4 & 80.2 & +45.72 & [44.24, 47.23] \\
\midrule
GPT-4o-mini & Standard RAG & 27.1 & 73.8 & +46.67 & [45.04, 48.32] \\
 & \hybridrag & 31.1 & 73.8 & +42.67 & [41.03, 44.23] \\
 & RAPTOR & 43.0 & 73.8 & +30.77 & [29.04, 32.53] \\
 & \textsc{MemGPT} & 50.2 & 73.8 & +23.57 & [22.00, 25.12] \\
 & \textsc{HippoRAG} & 31.0 & 73.8 & +42.73 & [41.16, 44.34] \\
\bottomrule
\end{tabular}
}
\caption{Paired bootstrap audit for the final 3,000-question main-table runs. The closest final API-backed comparison is Gemini~2.5 Flash with \textsc{MemGPT}; the improvement remains significant under paired resampling.}
\label{tab:paired-significance}
\end{table}

\begin{table}[!htbp]
\centering
\scriptsize
\setlength{\abovecaptionskip}{2pt}
\setlength{\belowcaptionskip}{0pt}
\setlength{\tabcolsep}{2.8pt}
\resizebox{\columnwidth}{!}{%
\begin{tabular}{lcccccc}
\toprule
Method & Domain Acc Std. & Batch Acc Range & Ctx Mean & Ctx Median & Ctx P95 & Ctx Max \\
\midrule
Standard RAG & 3.73 & 20.4--30.0 & 929.5 & 937 & 1076 & 1259 \\
\hybridrag & 3.65 & 25.0--43.2 & 1193.6 & 1194 & 1200 & 1200 \\
RAPTOR & 7.62 & 37.4--45.5 & 790.5 & 780 & 996 & 1219 \\
\textsc{MemGPT} & 7.70 & 52.6--72.7 & 2301.9 & 2317 & 2608 & 3359 \\
\textsc{HippoRAG} & 3.97 & 20.8--29.7 & 748.9 & 757 & 891 & 1043 \\
\textbf{\tsim} & 8.08 & 64.5--76.9 & 1060.8 & 1037 & 1408 & 1950 \\
\bottomrule
\end{tabular}
}
\caption{Variance and retrieved-context token audit for the final \answermodel main-table run. Context-token columns summarize selected answer context, not benchmark-side constructed length.}
\label{tab:variance-token-audit}
\end{table}

Token accounting separates constructed context length, a benchmark-side packing target estimated as \(\mathrm{tokens}\approx1.3\times\) whitespace words, from retrieved-context tokens, the compact prompt fragments passed after memory selection. We use constructed tokens only to build identical text packages, not to claim provider-native tokenizer parity across Gemma, OpenAI, Gemini, and DeepSeek.

\paragraph{Rationale similarity diagnostic.}
While Accuracy and CL Hit are primary, Table~\ref{tab:cross-backend-metrics} also reports rationale similarity as a secondary diagnostic. The evaluator parses each output into an answer choice and rationale, embeds the predicted rationale and gold \(q[\texttt{reasoning}]\) with SentenceTransformers \citep{reimers-gurevych-2019-sentence}, and computes cosine similarity. Reported runs use \texttt{BAAI/bge-large-en-v1.5}; \texttt{all-MiniLM-L6-v2} is retained only as a portability fallback. Missing rationales or evaluator failures receive \(0\), and table values average per-example scores over the completed full-dataset run. Rationale similarity is therefore a directional semantic-overlap diagnostic, not a cross-method ranking metric: a rationale can mention the same objects or policies without applying the binding local constraint, as illustrated by \tunedrag's \(0.487\) versus \tsim's \(0.485\) despite \(56.2\%\) versus \(73.8\%\) accuracy. This column is not used for ranking or calibration.

\paragraph{Reproducibility artifacts.}
\label{app:reproducibility}
The public repository provides the dataset card, protocol, validation report, exact-evidence audit, deterministic runtime builder, and \tsim reference implementation at \url{https://github.com/LordTARN1SHED/SCALE-QA}. Reproduction manifests document the local \answermodel 128k configuration and the \texttt{openai/gpt-4o-mini} full-dataset ablation settings.

\end{document}